\documentclass{article}

\usepackage{arxiv}
\usepackage[utf8]{inputenc} 
\usepackage[T1]{fontenc}    

\usepackage{algorithm}
\usepackage{algpseudocode}
\usepackage{amsmath}
\usepackage{amssymb}
\usepackage{graphicx}
\usepackage{booktabs}
\usepackage{tabularx}
\usepackage{array}
\usepackage{ragged2e}
\usepackage{makecell}
\usepackage{pifont}
\usepackage[table]{xcolor}
\usepackage[normalem]{ulem}
\usepackage{comment}
\usepackage{tikz}

\usepackage{pdflscape}
\usepackage{caption}

\usepackage{placeins}
\usepackage{float}

\usepackage{colortbl}   
\usepackage{subcaption}
\usepackage[export]{adjustbox} 

\usepackage{rotating}
\usepackage{multirow}

\definecolor{tickgreen}{RGB}{0,130,60}
\definecolor{crossred}{RGB}{180,35,35}

\providecommand{\rs}[2]{\mbox{$#1 \pm #2$}}
\newcommand{\rsb}[2]{\mbox{$\mathbf{#1} \pm \mathbf{#2}$}}
\newcommand{\rsu}[2]{\mbox{$\uline{#1 \pm #2}$}}

\newcolumntype{L}[1]{>{\raggedright\arraybackslash}p{#1}}
\newcolumntype{C}[1]{>{\centering\arraybackslash}p{#1}}

\definecolor{tabgray}{RGB}{247,247,247}
\definecolor{kipogray}{RGB}{235,248,240}

\newcommand{\heatmap}[2]{%
  \begin{subfigure}[t]{0.245\textwidth}
    \centering
    \adjustbox{width=\linewidth, keepaspectratio}{%
      \includegraphics{figs/heatmaps/#1.pdf}%
    }
    \caption{#2}
    \label{fig:hm:#1}
  \end{subfigure}%
}

\definecolor{rowgray}{gray}{0.93}

\scriptsize
\renewcommand{\arraystretch}{1.05}

\usepackage{hyperref}

\usepackage{url}            
\usepackage{amsfonts}       
\usepackage{nicefrac}       
\usepackage{microtype}      

\title{K-IPO: Kendall-constrained Importance Preserving Oversampling for Imbalanced Tabular Data}

\author{%
\parbox{0.96\textwidth}{%
\centering
\textbf{Marios Tyrovolas}\textsuperscript{1,2,3,*}
\quad
\textbf{Argiris Sofotasios}\textsuperscript{3,4}
\quad
\textbf{Dimitris Metaxakis}\textsuperscript{3,4}
\quad
\textbf{Georgios Mermigkis}\textsuperscript{2,3,4}
\\[0.55em]
\textbf{George Georgoulas}\textsuperscript{1,5}
\quad
\textbf{Panagiotis Hadjidoukas}\textsuperscript{2,4}
\quad
\textbf{Chrysostomos Stylios}\textsuperscript{1,2}
\\[0.9em]
{\normalfont\small
\textsuperscript{1}Department of Informatics and Telecommunications,
University of Ioannina, Arta, Greece
\\
\textsuperscript{2}Industrial Systems Institute,
Athena Research Center, Patras, Greece
\\
\textsuperscript{3}Archimedes Unit,
Athena Research Center, Athens, Greece
\\
\textsuperscript{4}Department of Computer Engineering and Informatics,
University of Patras, Patras, Greece
\\
\textsuperscript{5}Department of Mechanical Engineering and Aeronautics,
University of Patras, Patras, Greece
\\[0.45em]
\textsuperscript{*}\textit{Corresponding author:}
\href{mailto:tirovolas@kic.uoi.gr}
{\texttt{tirovolas@kic.uoi.gr}}
}%
}%
}

\begin{document}

\maketitle

\begin{abstract}
Oversampling is widely used to address class imbalance in tabular classification, but existing methods can distort the feature importance ranking underlying model explanations. Although recent studies have quantified this distortion by comparing real and synthetic data, none have actively sought to prevent it. In this paper, we introduce Kendall-constrained Importance-Preserving Oversampling (K-IPO), a generator-agnostic, "generate-then-select" framework that preserves the original data’s feature importance ranking during augmentation. K-IPO iteratively generates minority-class candidates and accepts them only if their inclusion maintains a user-defined minimum Kendall’s tau ($\tau$) correlation with the reference ranking. Optionally, stricter constraints can be applied to the highest-ranked features. We evaluated K-IPO on 20 imbalanced binary classification datasets using three classifiers and multiple explanation methods. In most cases, K-IPO achieved the best or tied-best results in feature importance preservation, explanation consistency, and class separability. It also generally improved predictive performance while maintaining competitive computational overhead.
\end{abstract}


\section{Introduction}
Class imbalance, in which observations are unevenly distributed across class labels, remains a significant challenge in supervised classification \cite{Chen2024SurveyImbalancedLearning}. Models trained on imbalanced data tend to favor majority-class patterns, resulting in skewed decision boundaries, reduced sensitivity to the minority class, and misleading performance estimates \cite{He2009LearningImbalancedData}. This issue is particularly critical when the minority class represents rare but high-impact events such as fraud, disease, safety incidents, or equipment failure \cite{QezelbashChamak2026KANBalance}, thereby limiting the real-world applicability of these models.

Common approaches for addressing class imbalance include data-level rebalancing, cost-sensitive or class-weighted learning, threshold adjustment, ensemble methods, and hybrid strategies \cite{He2009LearningImbalancedData, Elkan2001CostSensitiveLearning, Zhou2006TrainingCostSensitiveNNs, Liu2009ExploratoryUndersampling, Zhao2025ContributionHybridResampling}. However, the latter three typically require task- or classifier-specific cost calibration or introduce additional training and tuning complexity. Therefore, data-level rebalancing remains a widely adopted strategy, with oversampling being a practical option because it is model-agnostic, easy to integrate into standard training pipelines, and does not require modifications to the downstream classifier. Accordingly, numerous oversampling methods have been proposed, ranging from random replication \cite{Batista2004BalancingTrainingData} and local interpolation \cite{Chawla2002SMOTE} to probabilistic \cite{Aich2025CopulaSMOTE} and deep-generative approaches \cite{Xu2019CTGAN, Kotelnikov2023TabDDPM, Schultz2024ConvGeN}. Although these methods can improve class balance and predictive performance, they are designed primarily to optimize sample placement, distributional fidelity, or downstream utility rather than explicitly preserving the feature-target relationships that shape model reasoning \cite{Hansen2023SyntheticTabularDataCentricAI}.

This limitation becomes particularly important when evaluation considers not only predictive performance, but also whether a classifier relies on appropriate feature-target relationships rather than merely achieving accurate predictions \cite{Ross2017RightReasons}. Synthetic samples may improve minority-class metrics while altering which predictors appear to be the most influential to the trained model. Previous studies have shown that synthetic tabular data may fail to preserve deeper structural properties, including logical or functional dependencies and feature importance rankings \cite{Umesh2025DependenciesSyntheticTabularData, Giles2022FakingFeatureImportance, Leidiyana2026SMOTEFeatureImportance}. Such discrepancies have direct implications for eXplainable Artificial Intelligence (XAI), where feature importance serves as a primary tool for model interpretation, auditing, domain validation, and decision support \cite{Goldwasser2025FeatureImportanceRankings}. If oversampling modifies the ranking of influential predictors, a model may remain accurate while producing explanations that no longer reflect the original prediction task. In high-stakes domains such as healthcare, manufacturing, and finance, this explanation drift can undermine trust, reproducibility, and practical reliability.

Recent frameworks have recognized the need to evaluate synthetic data beyond predictive utility by incorporating structural fidelity and explanation-aware criteria \cite{Yu2025SHAPDistance, Kapar2026SyntheticTabularDataXAI}. However, existing approaches remain primarily diagnostic, measuring feature importance drift after synthetic data generation rather than preventing it during augmentation. This leaves a methodological gap: oversampling should not only generate plausible and useful minority-class samples but also preserve the feature importance structure derived from the original data.

To address this gap, we propose Kendall-constrained Importance-Preserving Oversampling (K-IPO), a "generate-then-select" framework for imbalanced tabular classification. K-IPO employs a chosen oversampling method or tabular generator to iteratively produce minority-class candidates, accepting only candidate sets whose inclusion preserves the feature importance ranking of the original data. Rank preservation is enforced through Kendall’s rank correlation coefficient, $\tau$ \cite{Kendall1938RankCorrelation}, with optional constraints applied to the top-$k$ features. Thus, K-IPO integrates explanation stability directly into the oversampling process rather than assessing it only post hoc. Using 20 imbalanced binary tabular classification datasets, we evaluate K-IPO in terms of feature-importance ranking preservation and class separability of the generated data, as well as the predictive performance and explanation consistency of classifiers trained on the augmented data. Comparisons with representative baselines show that K-IPO achieves the strongest aggregate predictive–explanatory performance while incurring acceptable additional computational overhead.

The remainder of this paper is organized as follows. Section~\ref{literature_review} reviews related work on imbalanced learning, tabular data generation, and explanation-aware evaluation. Section~\ref{methodology} presents K-IPO, while Section~\ref{experimental_setup} describes the experimental setup, including datasets, preprocessing, baseline oversampling methods, classifiers, explainability methods, and evaluation metrics. Section~\ref{results} presents the empirical results, and Section~\ref{conclusions} concludes the paper by discussing the limitations of the proposed framework and outlining future research directions.

\section{Literature Review}
\label{literature_review}

This section reviews oversampling methods for imbalanced tabular data classification from the perspective of feature importance preservation. Existing methods can be broadly categorized into \textit{replication-based oversampling}, \textit{local interpolation}, \textit{distribution-learning approaches}, and \textit{deep generative models}. Although these categories differ in how they generate minority-class observations, they generally prioritize class balance, local geometry, distributional fidelity, or downstream predictive utility over preserving the feature-target relationships underlying model explanations.

\subsection{Replication-based and interpolation-based oversampling}

Random oversampling is the simplest data-level approach to address class imbalance, achieved by duplicating minority class observations until the desired class ratio is reached. Although this method increases minority-class representation, it introduces no new information about the minority distribution and may lead to overfitting when few distinct minority observations are available \cite{Batista2004BalancingTrainingData, He2009LearningImbalancedData}. This limitation motivated the development of synthetic approaches, such as the Synthetic Minority Oversampling Technique (SMOTE) \cite{Chawla2002SMOTE}. SMOTE generates minority-class samples by interpolating between an observation and its nearest minority-class neighbors, making it classifier-independent and easy to integrate into standard learning pipelines.

Despite its success, SMOTE assumes that local linear interpolation between minority class observations produces plausible samples. This assumption may fail when minority regions are sparse, multimodal, noisy, or overlapping with the majority class. In such cases, synthetic samples may connect distinct minority subclusters, populate ambiguous regions, or smooth the minority distribution in ways that distort the original learning problem \cite{Elreedy2024SMOTEDistributionAnalysis}. In response, several SMOTE variants have been proposed to refine the generation or selection process. For instance, Borderline-SMOTE focuses on minority observations near the decision boundary \cite{Han2005BorderlineSMOTE}, ADASYN allocates more synthetic samples to minority observations that are harder to learn \cite{He2008ADASYN}, and hybrid approaches, such as SMOTE-Tomek and SMOTE-ENN, combine oversampling with cleaning mechanisms that remove ambiguous, noisy, or borderline observations \cite{Batista2004BalancingTrainingData}. SMOTE with Boosting (SMOTEWB) similarly combines noise detection with boosting-guided generation~\cite{Saglam2022SMOTEWB}. Specifically, it uses boosting-derived weights to determine where samples should be generated and adaptively selects the number of neighbors for each observation. Clustering-based methods, such as KMeans-SMOTE, restrict generation to selected minority regions to reduce noise and address within-class imbalance \cite{Douzas2018KMeansSMOTE}. Nevertheless, these approaches remain primarily guided by geometric, neighborhood-based, or heuristic criteria and do not assess whether synthetic observations preserve the global feature importance ranking derived from the original training data.

\subsection{Distribution-learning methods for tabular synthesis}

Distribution learning methods explicitly model the minority class distribution. Instead of interpolating between neighboring observations, they estimate the statistical properties of the data and sample new records from the learned distribution. Copula-based methods are a prominent example because they separate marginal distributions from the dependence structure, allowing heterogeneous variables to be modeled more flexibly than methods based solely on Euclidean neighborhoods \cite{Aich2025CopulaSMOTE}. This flexibility is particularly valuable in tabular data, where variables may differ in scale, type, and dependency structure. Nevertheless, distribution learning remains challenging under class imbalance because the limited number of minority observations complicates the reliable estimation of marginal distributions and dependence structures. Common copula formulations may also struggle to capture nonlinear dependencies, higher-order interactions, and minority-specific structures \cite{Feldman2024NonparametricCopulaModels, Huk2025CopulaClassifier}. Consequently, these methods may reproduce some statistical regularities of the minority class without preserving the interactions that shape model behavior and feature attribution.

\subsection{Deep generative oversampling and tabular diffusion models}

Deep generative models have gained prominence in generating synthetic tabular data due to their ability to capture nonlinear dependencies, handle mixed continuous and categorical variables, and model complex feature interactions. Generative Adversarial Networks (GANs), Variational Autoencoders (VAEs), and diffusion models have all been employed for this purpose \cite{Shi2025SyntheticTabularDataSurvey}.

CTGAN and TVAE are widely used deep learning baselines for tabular synthesis. CTGAN adapts the GAN framework to address two common properties of real-world tabular data: non-Gaussian or multimodal continuous variables and highly imbalanced categorical variables. It combines mode-specific normalization for continuous columns with conditional generation and training-by-sampling for discrete columns, improving the representation of both frequent and rare categories \cite{Xu2019CTGAN}. Introduced in the same study, TVAE provides a VAE-based alternative for mixed-type tabular data. Instead of using adversarial training, it learns a latent probabilistic representation of the observed rows and generates new records by decoding samples from this space \cite{Xu2019CTGAN}. TabDDPM extends denoising diffusion probabilistic models to tabular data by formulating generation as a gradual denoising process adapted to heterogeneous feature spaces containing continuous and discrete variables. Therefore, it represents an important modern baseline connecting tabular synthesis with the broader success of diffusion-based generative modeling \cite{Kotelnikov2023TabDDPM}.  

Although these models are powerful candidates for oversampling, their objectives do not guarantee feature importance preservation. GANs optimize adversarial indistinguishability, VAEs emphasize reconstruction and latent-likelihood objectives, and diffusion models optimize denoising. Thus, even realistic synthetic samples that improve downstream performance may weaken relevant dependencies, amplify spurious associations, or alter the ranking of influential predictors. This risk is especially pronounced under class imbalance, where minority-class structure must be inferred from limited, potentially sparse, noisy, or overlapping observations. Empirical studies further indicate that the benefits of deep generative oversampling may be modest relative to its complexity and strongly dependent on dataset characteristics \cite{Camino2020OversamplingTabularData, Engelmann2021CWGANOversampling, Dsouza2025OverlapClass}. Deep generative models are therefore valuable candidate generators, but preventing feature importance drift requires an additional acceptance criterion.

\begin{table*}[t]
\centering
\footnotesize
\setlength{\tabcolsep}{5.0pt}
\renewcommand{\arraystretch}{1.20}

\newcommand{\NoCircle}{%
  \tikz[baseline=-0.6ex]\draw[line width=0.5pt] (0,0) circle (0.75ex);%
}
\newcommand{\YesCircle}{%
  \tikz[baseline=-0.6ex]\fill (0,0) circle (0.75ex);%
}

\caption{Conceptual positioning of oversampling families for imbalanced tabular classification.}
\label{tab:method_positioning}

\begin{tabularx}{\textwidth}{@{}
>{\raggedright\arraybackslash}p{0.20\textwidth}
>{\raggedright\arraybackslash}p{0.27\textwidth}
>{\raggedright\arraybackslash}X
>{\centering\arraybackslash}p{0.17\textwidth}
@{}}
\toprule
\textbf{Family}
&
\textbf{Representative methods}
&
\textbf{Generation principle}
&
\textbf{FI-rank preservation} \\
\midrule

\rowcolor{tabgray}
Replication-based
& Random oversampling
& Duplicates existing minority observations.
& \NoCircle \\

Local interpolation
& SMOTE, SMOTE-NC; Borderline-SMOTE, ADASYN; SMOTE-Tomek, SMOTE-ENN; KMeans-SMOTE
& Interpolates between neighboring minority observations.
& \NoCircle \\

\rowcolor{tabgray}
Distribution-learning
& Gaussian Copula
& Samples from an estimated minority-class distribution.
& \NoCircle \\

Deep generative
& CTGAN, TVAE
& Learns a neural generator for tabular records.
& \NoCircle \\

\rowcolor{tabgray}
Diffusion-based
& TabDDPM
& Generates records through iterative denoising.
& \NoCircle \\

\midrule
\textbf{Proposed method}
& \textbf{K-IPO}
& \textbf{Generates candidate samples and accepts them only under a Kendall-based feature importance (FI) ranking constraint.}
& \YesCircle \\

\bottomrule
\end{tabularx}
\end{table*}

\subsection{Synthetic-data evaluation and feature importance preservation}

Synthetic tabular data are commonly evaluated in terms of fidelity, utility, and privacy \cite{Platzer2021MixedTypeSyntheticData, Lautrup2025Syntheval, Hernandez2025SyntheticTabularHealthEvaluation}. Fidelity metrics assess whether synthetic data reproduce properties of the original data, including marginal distributions, correlations, and multivariate similarities. Utility metrics evaluate their suitability for downstream tasks, often through predictive performance comparisons or train-on-synthetic-test-on-real protocols. Privacy metrics estimate whether synthetic records disclose information about individuals in the original data using approaches such as nearest-neighbor and inference-risk analyses. These criteria are complementary rather than interchangeable: a synthetic dataset may support accurate predictions while failing to preserve structural relationships required for interpretation.

Recent research has therefore emphasized structure- and explanation-aware evaluation. Giles \textit{et al.}~\cite{Giles2022FakingFeatureImportance} showed that feature importance rankings derived from synthetic data can differ from those obtained from real data, particularly in complex real-world scenarios. Umesh \textit{et al.}~\cite{Umesh2025DependenciesSyntheticTabularData} similarly found that synthetic tabular generators may fail to preserve the logical and functional dependencies among features. In imbalanced classification, Stando \textit{et al.}~\cite{Stando2024BalancingMethodsModelBehavior} demonstrated that balancing methods can alter model behavior in ways not captured by conventional performance metrics. They compared models before and after rebalancing using variable importance, partial dependence, and accumulated local effects. Explanation-aware measures, such as SHapley Additive exPlanations (SHAP)-based Distance, have also shown that attribution patterns can reveal semantic discrepancies between real and synthetic data that distributional and predictive metrics may overlook \cite{Yu2025SHAPDistance}.

\section{K-IPO Framework}
\label{methodology}

\begin{figure}
    \centering
    \includegraphics[width=1\linewidth]{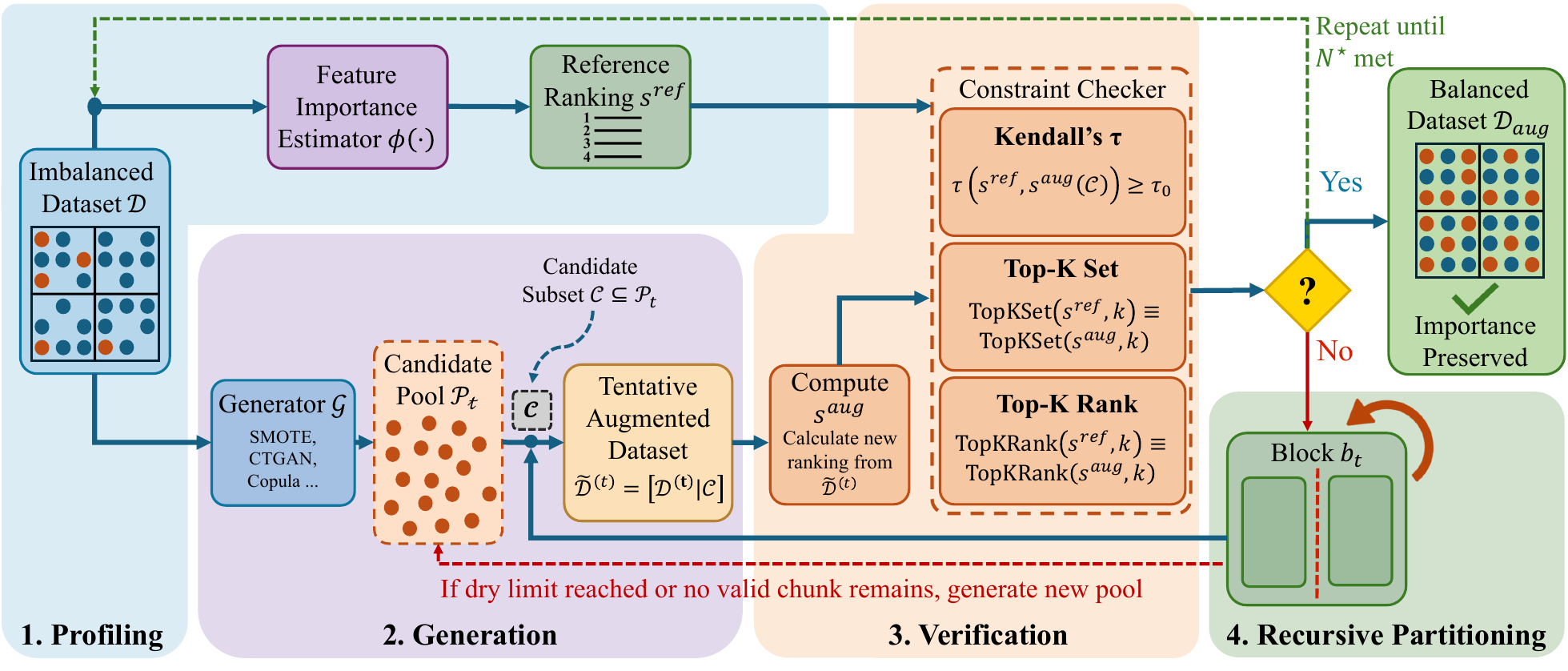}
    \caption{K-IPO workflow for importance-preserving oversampling. The method profiles the original data to compute a reference feature importance ranking, generates minority-class candidates, verifies their effect on the ranking using Kendall's $\tau$ and optional top-$k$ constraints, and recursively partitions rejected blocks until $N^\star$ valid synthetic samples are accepted.}
    \label{fig:kipo_workflow}
\end{figure}

This section presents Kendall-constrained Importance-Preserving Oversampling (K-IPO), a framework designed for oversampling in imbalanced tabular classification that explicitly mitigates feature importance drift during data augmentation. Unlike the conventional methods discussed in Section~\ref{literature_review}, which primarily focus on balancing the training distribution, K-IPO accepts synthetic samples only when they preserve the feature importance ranking of the original training data within a user-defined tolerance (see Table~\ref{tab:method_positioning}). Fig.~\ref{fig:kipo_workflow} illustrates the overall workflow, while Algorithm~\ref{alg:kipo_short} summarizes the oversampling procedure of the proposed framework. The following subsections describe the oversampling objective, generator-agnostic candidate generation, acceptance mechanism, adaptive block-and-chunk search, and computational requirements.

\subsection{Problem Formulation}

Let $\mathcal{D} =\{(x_i,y_i)\}_{i=1}^{n}$ be an imbalanced binary
tabular dataset, where $x_i\in\mathbb{R}^{d}$ is a $d$-dimensional feature vector, and $y_i\in\{0,1\}$ is its class label. Without loss of generality, $y=1$ denotes the minority class, and $y=0$ denotes the majority class. Additionally, let $n_{maj}$ and $n_{min}$ denote the numbers of majority and minority class samples, respectively. For a target minority-to-majority ratio $r>0$, the objective is to construct an augmented dataset containing approximately $r \cdot n_{maj}$ minority-class observations. Therefore, the required number of synthetic samples is
\begin{equation}
    N^\star=\max\left(0,\left\lfloor r \cdot n_{maj}\right\rfloor-n_{min}\right).
\end{equation}

K-IPO aims to achieve this target while preserving the feature-importance structure of the original training data. Let $s^{ref} = \phi(\mathcal{D})$ denote the reference importance-score vector produced by an importance estimator $\phi(\cdot)$. The ranking induced by $\mathbf{s}^{ref}$ is used as the reference order throughout oversampling. The framework is agnostic to the choice of $\phi$, which may be a filter-based estimator such as the ANOVA $F$-score~\cite{Fisher1992StatisticalMethodsResearchWorkers}, or an XAI method such as SHAP~\cite{Lundberg2017SHAP} or Permutation Feature Importance (PFI)~\cite{Breiman2001RandomForests}. In this study, $\phi$ was instantiated using the ANOVA $F$-score, which assigns larger values to features with greater between-class variation relative to within-class variation. This choice was made because it is model-independent, computationally efficient, and suitable for repeated evaluation during iterative oversampling~\cite{Elkhawaga2024TrustExplanationXAI}.

\subsection{Generator-Agnostic Candidate Pool Generation}

At iteration $t$, let $\mathcal{D}^{(t)}$ denote the current augmented dataset and let $N_{acc}$ be the number of synthetic samples accepted so far. K-IPO requests a candidate block of size 
\begin{equation}
    b_t=\min\left(B_{\max},N^\star-N_{acc}\right)
\end{equation}
and generates
\begin{equation}
    \mathcal{P}_t
    =
    \mathcal{G}\left(\mathcal{D}^{(t)},b_t\right),
    \qquad |\mathcal{P}_t|=b_t.
\end{equation}

The generator $\mathcal{G}$ is a modular component that can be instantiated using conventional oversamplers, such as SMOTE or ADASYN, probabilistic tabular synthesizers, or deep generative models. K-IPO does not modify the generator itself; instead, it applies an importance-preserving selection layer to the generated candidate samples. As a result, the data generation method can be chosen based on the data, modelling constraints, and application requirements without changing the acceptance mechanism.

\subsection{Importance-Preserving Acceptance Mechanism}

For a candidate subset $\mathcal{C}\subseteq\mathcal{P}_t$, K-IPO forms
the tentative dataset
\begin{equation}
    \widetilde{\mathcal{D}}^{(t)}
    = \mathbf{[\mathcal{D}^{(t)}|\:\mathcal{C}]},
\end{equation}
where $\mathbf{[\cdot|\cdot]}$ denotes row-wise dataset concatenation, and recomputes the importance scores as
\begin{equation}
    \mathbf{s}^{aug}(\mathcal{C})
    =
    \phi\left(\widetilde{\mathcal{D}}^{(t)}\right).
\end{equation}

The subset is accepted only if the ranking induced by $s^{aug}(\mathcal{C})$ is sufficiently aligned with the reference ranking. The primary acceptance criterion is Kendall's rank correlation coefficient:
\begin{equation}\tau\left(s^{ref},s^{aug}\left(\mathcal{C}\right)\right)\geq\tau_0,
\label{eq:kipo_acceptance}
\end{equation}
where $\tau_0\in\left(0,1\right]$ is a user-defined minimum permitted Kendall rank correlation. Kendall's $\tau$ measures ordinal agreement by comparing concordant and discordant feature pairs \cite{Kendall1938RankCorrelation}. Values close to $1$ indicate strong preservation of the reference ordering, whereas lower values indicate increasing rank drift. In addition to this global criterion, K-IPO can optionally enforce stricter local consistency through user-defined top-$k$ constraints. Specifically, the top-$k$ overlap constraint requires that the set of the $k$ most important features remains unchanged, while the top-$k$ ordering constraint requires that the internal ordering of these features also remains unchanged:
\begin{equation}
    \mathrm{TopKSet}(s^{ref},k) \equiv \mathrm{TopKSet}(s^{aug},k), \quad \mathrm{TopKRank}(s^{ref},k) \equiv \mathrm{TopKRank}(s^{aug},k).
\end{equation}

A third, more restrictive option limits changes in the magnitudes of the relative importance scores of the top-ranked features. Because such a constraint is generally suitable only for small datasets with specific score characteristics, all three top-$k$ criteria are optional and user-defined. Practitioners may enforce overlap alone, overlap and ordering, or an additional magnitude constraint. The K-IPO configuration evaluated in this study did not use magnitude preservation and was therefore omitted from Fig.~\ref{fig:kipo_workflow}.

\subsection{Adaptive Block-and-Chunk Search Strategy}

K-IPO first evaluates the complete candidate block $\mathcal{P}_t$. If the block satisfies Eq.~\eqref{eq:kipo_acceptance} and any enabled top-$k$ constraints, it is accepted in full. Otherwise, rejecting the entire block may be overly conservative when only a subset of samples violates the acceptance criteria. At the same time, evaluating samples individually can increase selection bias and require substantially more feature importance computations.

To balance these concerns, K-IPO uses an adaptive block-and-chunk search. It initializes the chunk size as $c=\lfloor|\mathcal{P}t|/2\rfloor$,  partitions the remaining candidates into contiguous chunks of size $c$, and halves $c$ after each pass until $c<B{\min}$. Chunks are evaluated sequentially against the current augmented dataset. Each accepted chunk is immediately added to the dataset and removed from the candidate pool, ensuring that subsequent evaluations account for all samples accepted earlier in the iteration. In this way, every dataset update satisfies the preservation criteria.

The search is terminated and a new candidate pool is generated if no acceptable sub-block is found before $c$ falls below $B_{\min}$, or if the number of consecutive chunk rejections reaches the maximum dry-attempt limit. This strategy improves candidate utilization and robustness to imperfect generators while balancing feature importance preservation with computational cost.

\begin{algorithm}[t]
\caption{Kendall-constrained Importance-Preserving Oversampling (K-IPO)}
\label{alg:kipo_short}
\begin{algorithmic}[1]
\Require Dataset $\mathcal{D}$, target ratio $r$, generator $\mathcal{G}$, importance function $\phi$, Kendall threshold $\tau_0$, optional top-$k$ constraints, maximum block size $B_{\max}$, minimum block size $B_{\min}$, maximum dry attempts $A_{\max}$
\Ensure Balanced dataset $\mathcal{D}_{aug}$

\State Compute $N^\star = \max(0,\lfloor r \cdot n_{maj}\rfloor - n_{min})$,
$N_{acc}\gets0$
\State $\mathcal{D}_{aug}\gets\mathcal{D}$,  $\mathbf{s}^{ref}\gets\phi(\mathcal{D})$
\While{$N_{acc} < N^\star$}
    \State $b_t \gets \min(B_{\max}, N^\star-N_{acc})$
    \State Generate candidate pool $\mathcal{P}_{t} \gets \mathcal{G}(\mathcal{D}^{(t)} , b_t)$

    \If{\Call{ImportancePreservingAccept}{$\mathcal{P}_{t}, \mathcal{D}_{aug}, \mathbf{s}^{ref}$}}
        \State $\mathcal{D}_{aug} \gets \mathbf{[ \mathcal{D}_{aug}|\mathcal{P}_{t}]} $

        \State $N_{acc} \gets N_{acc} +  |\mathcal{P}_{t}|$
    \Else
        \State $\mathcal{R} \gets \mathcal{P}_{t}$,
        $c\gets\lfloor|\mathcal{R}|/2\rfloor$, $dry\gets0$

        \While{$c\geq B_{\min}$ \textbf{and} $dry<A_{\max}$}
            \State Partition $\mathcal{R}$ into contiguous chunks of size $c$
           \If{$\mathrm{mod}(|R|, c) \neq 0$}
                \State Discard last chunk
            \EndIf

            \For{each evaluated chunk $\mathcal{C}$}
                \If{\Call{ImportancePreservingAccept}{$\mathcal{C}, \mathcal{D}_{aug}, \mathbf{s}^{ref}$}}
                    \State $\mathcal{D}_{aug} \gets \mathbf{[ \mathcal{D}_{aug}|\mathcal{C}]}$
                    \State $N_{acc} \gets N_{acc} + |\mathcal{C}|$
                    \State $\mathcal{R}\gets\mathcal{R}\setminus\mathcal{C}$
                    \Comment{Remove accepted chunk}
                    \State $dry \gets 0$
                \Else
                    \State $dry \gets dry + 1$
                    \If{$dry \geq A_{\max}$}
                        \State \textbf{break} \Comment{Abandon current block search}
                    \EndIf
                \EndIf
            \EndFor
            \State $c\gets\lfloor c/2\rfloor$
        \EndWhile
    \EndIf
\EndWhile

\State \Return $\mathcal{D}_{aug} \gets \mathcal{D}^{(t)}$

\Function{ImportancePreservingAccept}{$\mathcal{C}, \mathcal{D}^{(t)}, \mathbf{s}^{ref}$}
    \State Compute $\mathbf{s}^{aug} = \phi(\mathbf{[ \mathcal{D}^{(t)}|\mathcal{C}]})$
    \State Accept if:
    \Statex \hspace{\algorithmicindent}
    $\tau(\mathbf{s}^{ref}, \mathbf{s}^{aug}) \geq \tau_0 \land\ \text{enabled top-}k\text{ constraints are satisfied}$
\EndFunction
\end{algorithmic}
\end{algorithm}

\subsection{Computational Requirements}
Two factors primarily determine the computational overhead of K-IPO: (i) the number of times the underlying generator produces candidate sample pools, and (ii) the number of times the importance-preserving acceptance mechanism is invoked.

Let $G$ denote the number of calls to the underlying generator, and let $M$ denote the number of calls to the acceptance function after the reference feature ranking has been computed. Furthermore, let $C_g(n,d,k)$ denote the cost of generating a candidate pool of $k$ samples from a dataset with $n$ samples and $d$ features, and let $C_{\phi}(n,d)$ denote the cost of computing the feature importance ranking on $n$ samples and $d$ features. The additional computational cost introduced by K-IPO can therefore be approximated as $G \cdot C_g(n,d,k) + M \cdot C_{\phi}(n,d)$, plus lower-order costs associated with ranking comparison, Kendall's $\tau$ computation, and optional top-$k$ consistency checks.

When most generated candidate blocks are accepted without further partitioning, the number of pool generation steps is significantly reduced, and the number of acceptance evaluations is approximately given by

\begin{equation}
 M \approx \left\lceil \frac{N^\star}{B_{\max}} \right\rceil.
\end{equation}

In this favorable case, the method benefits from block-level screening, requiring only a limited number of feature importance recomputations. However, when candidate blocks are repeatedly rejected and recursively split into smaller sub-blocks, the number of acceptance evaluations increases until either the minimum block size $B_{\min}$ is reached or the dry-attempt limit is exceeded. With the ANOVA $F$-score used in this study, $C_{\phi}(n,d)$ increases linearly with the number of samples in the dataset. Consequently, the overall runtime of K-IPO is mainly governed by the dataset scale, the block-size hyperparameters, the quality and computational complexity of the candidate generator, and the strictness of the Kendall's $\tau$ and optional top-$k$ preservation constraints. These factors induce a trade-off between computational efficiency and the degree of feature importance preservation.

\section{Experimental Setup}
\label{experimental_setup}

This section describes the experimental protocol adopted to evaluate the proposed framework, including the datasets, preprocessing steps, baseline oversampling methods, predictive models, XAI techniques, and evaluation metrics. All experiments were conducted using Python 3.10.20 on a workstation equipped with an AMD Ryzen 9 7950x processor, 32 GB of RAM, an NVIDIA GeForce RTX 4070 Ti GPU with 12 GB VRAM and a Kingston KC3000 PCIe 4.0 NVMe SSD. The source code and all experimental results are publicly available on GitHub\footnote{\url{https://github.com/CEID-HPCLAB/K-IPO}}.

\subsection{Datasets} \label{datasets}

We evaluated K-IPO on 20 tabular binary classification datasets from widely used repositories and collections, including UCI, OpenML, Kaggle, and imbalanced-learn, as well as domain-specific datasets from prior studies~\cite{Tyrovolas2024InformationFlowFCM, Parginos2025PowerGridOverloadIFFCM}. The benchmark spans diverse application domains, sample sizes, feature dimensionalities and types, and class-imbalance ratios, including numerical, categorical, and mixed-type data. For seven datasets whose original class distributions were either balanced or only mildly imbalanced, namely airlines, bank-customer-churn-prediction, churn, rl, fried, bank32nh, and online shoppers purchasing intention, we artificially induced imbalance using imbalanced-learn's \texttt{make\_imbalance()} function, setting the minority-to-majority ratio to approximately 0.1.

For each dataset, we extracted the target variable and encoded the majority and minority classes as 0 and 1, respectively. Categorical features were label- or ordinal-encoded as appropriate, while numerical features were standardized using Z-score normalization. All transformations were fitted on the training data and applied to the corresponding test split. Table~\ref{tab:datasets} summarizes the sample size, feature composition, imbalance ratio, and separability score, $1-N3$, where $N3$ is the leave-one-out error rate of a 1-nearest-neighbor classifier \cite{Ho2002ComplexityMeasures}. Higher $1-N3$ values indicate greater local class separability and, therefore, lower classification complexity.

\begin{table}[!htbp]
\centering
\caption{Detailed description of binary imbalanced datasets used in the experiments.}
\label{tab:datasets}

\resizebox{0.75\textwidth}{!}{%
\begin{tabular}{llccccccc}
\toprule
ID & Dataset & \#$N$ & \#$F_n$ & \#$F_c$ & Maj. & Min. & IR & $1-N3$ \\
\midrule
D1  & ai4i 2020                  & 9981  & 5 & 1 & 9643 & 338  & 0.035 & 0.967 \\
D2  & abalone                    & 4177  & 7 & 1 & 3786 & 391  & 0.103 & 0.866 \\
D3  & airlines                   & 16427 & 3 & 4 & 14934 & 1493 & 0.100 & 0.842 \\
D4  & seismic-bumps              & 2584  & 11 & 4 & 2414 & 170  & 0.070 & 0.896 \\
D5  & bank customer churn prediction        & 8759 & 8 & 2 & 7963 & 796  & 0.100 & 0.891 \\
D6 & churn                      & 4722  & 17 & 3 & 4293 & 429  & 0.100 & 0.903 \\
D7  & bank-marketing             & 4521  & 7 & 9 & 4000 & 521  & 0.130 & 0.872 \\
D8 & rl                         & 2733  & 5 & 7 & 2485 & 248  & 0.100 & 0.867 \\
D9 & online shoppers purchasing intention            & 11464 & 15 & 2 & 10422 & 1042 & 0.100 & 0.897 \\
D10  & car\_eval\_4             & 1728 & 0 & 6 & 1663 & 65  & 0.039 & 0.965 \\
D11  & mammography                & 11183 & 6 & 0 & 10923 & 260  & 0.024 & 0.984 \\
D12  & ur3 cobotops               & 7355  & 20 & 0 & 6837 & 518  & 0.076 & 0.925 \\
D13 & nhanes                     & 2105  & 7 & 0 & 1914 & 191  & 0.100 & 0.847 \\
D14 & magic gamma telescope      & 13565 & 10 & 0 & 12332 & 1233 & 0.100 & 0.925 \\
D15 & fried                      & 22469 & 10 & 0 & 20427 & 2042 & 0.100 & 0.924 \\
D16 & bank32nh                   & 6213  & 32 & 0 & 5649 & 564  & 0.100 & 0.877 \\
D17 & wilt                       & 4839  & 5 & 0 & 4578 & 261  & 0.057 & 0.971 \\
D18 & lines-overload-50          & 4368  & 187 & 0 & 4046 & 322  & 0.080 & 0.978 \\
D19 & japanese vowels            & 9961  & 14 & 0 & 8347 & 1614 & 0.193 & 0.998 \\
D20 & pen\_digits                 & 10992 & 16 & 0 & 9937 & 1055 & 0.106 & 0.999 \\
\bottomrule
\end{tabular}
} 

\vspace{0.2em}

\begin{minipage}{0.75\textwidth}
\footnotesize
\textit{Note:} \#$N$ denotes the number of samples; \#$F_n$ and \#$F_c$ denote the numbers of numerical and categorical features, respectively; Maj. and Min. denote the numbers of samples in the majority and minority classes, respectively; IR denotes the minority-to-majority class ratio, calculated as $\mathrm{Min.}/\mathrm{Maj.}$; and $1-N3$ is a class-separability indicator based on nearest-neighbor misclassification.
\end{minipage}
\end{table}

\subsection{Baseline oversampling and tabular generation methods}
\label{baseline_methods}

K-IPO was evaluated against representative baselines spanning the main oversampling and tabular generation paradigms reviewed in Section~\ref{literature_review}:

\begin{enumerate}
\item \textbf{SMOTE-family neighborhood-based oversampling}.
The SMOTE variant was selected a priori based on the feature types of each dataset. Mixed numerical-categorical datasets used SMOTE-NC~\cite{Chawla2002SMOTE}, which interpolates continuous features and assigns valid categorical values from the local minority neighborhood. The fully categorical car-eval-4 dataset used SMOTEN~\cite{Chawla2002SMOTE}, which defines neighborhoods using categorical distance and generates nominal values without interpolating the numerical encodings. Numerical-only datasets used SMOTEWB~\cite{Saglam2022SMOTEWB}, a recent noise-aware variant whose original evaluation reported significant MCC improvements over the standard SMOTE and random oversampling. Its competitiveness is further supported by a recent study of 18 imbalanced datasets, in which it outperformed established variants, including Borderline-SMOTE, ADASYN, SVM-SMOTE, ProWSyn, and polyfit-Mesh, in terms of the average F1-score rank with an SVM classifier~\cite{Kumbhar2026BayesianOversampling}.

\item \textbf{Statistical distribution learning}.
Gaussian Copula was included as a representative statistical generator. For a fixed training set, it fits the marginal distribution of each feature, models the dependencies among the transformed variables, and samples from the resulting joint distribution.

\item \textbf{GAN/VAE-based deep generative models}.
CTGAN and TVAE were selected as representative neural tabular generators. Widely used as benchmarks for adversarial and variational generation, respectively, both are designed for mixed-type tabular data.

\item \textbf{Diffusion-based generation}.
TabDDPM was included as a state-of-the-art denoising diffusion model for heterogeneous tabular data generation~\cite{Kotelnikov2023TabDDPM}.
\end{enumerate}

This selection enabled a comprehensive assessment of K-IPO against both traditional oversampling techniques and modern generative approaches~\cite{Chen2024SurveyImbalancedLearning}. For reproducibility and fair comparison across heterogeneous methods, all the baselines were evaluated using the default configurations recommended by their respective libraries.

\subsection{Predictive Models}

To mitigate the risk of biased results from using a single predictive model, all oversampling and tabular generation methods were evaluated using three classifiers of varying complexity and interpretability.
\begin{enumerate}
    \item \textbf{Random Forest}. Random Forest serves as a robust ensemble baseline for tabular classification. By aggregating multiple decision trees trained on random subsets of samples and features, it reduces variance, enhances robustness against overfitting, and captures nonlinear feature interactions, making it a standard benchmark for tabular prediction tasks.

    \item \textbf{Multi-Layer Perceptron (MLP)}. An MLP was included as a black-box neural classifier. It consists of stacked hidden layers with nonlinear activation functions and can learn complex hierarchical mappings between the input features and class labels.

    \item \textbf{GAMI-Net}. GAMI-Net was used as an explainable neural model based on disentangled additive subnetworks, where each subnetwork models either a single-feature effect or a pairwise interaction \cite{Yang2021GAMINet}. By first learning the main effects and then modeling interactions through residuals, GAMI-Net enables the inherent attribution of feature contributions.
\end{enumerate}

For these models, standard configuration parameters were used: 100 trees for Random Forest, default scikit-learn settings for the MLP, and default PiML configuration with available CPU parallelism for GAMI-Net~\cite{Sudjianto2023PiMLToolbox}. This ensured that the performance differences could be attributed primarily to the oversampling methods rather than classifier-specific optimization. Each classifier was trained independently on each augmented training set, and the final dataset-level results were obtained by averaging the performances across the three classifiers.

\subsection{XAI Methods and Feature Importance Extraction}

The objective of this study was not only to improve predictive performance but also to preserve the feature importance ranking of the original data. Therefore, each classifier was paired with an explanation method suitable for extracting feature importance.

For the Random Forest classifier, we employed \textbf{SHAP}, a game-theoretic post-hoc explanation framework that quantifies each feature's contribution using Shapley values, representing its average marginal effect across all possible feature subsets~\cite{Lundberg2017SHAP}. Global feature importance rankings were obtained using TreeSHAP, with importance scores computed as the mean absolute SHAP values across test instances.

For the MLP, feature importance was estimated using \textbf{Sparseness-Optimized Feature Importance (SOFI)}, a post-hoc method that measures the impact of controlled feature marginalization on predictive performance~\cite{Grau2024SparsenessOptimizedFeatureImportance}. Prior work has shown that SOFI produces more faithful and accurate feature rankings than alternative post-hoc methods, as indicated by the stronger correspondence between feature removal and performance degradation.

For \textbf{GAMI-Net}, feature importance was extracted directly from the fitted model through its intrinsic interpretability, eliminating the need for a separate explanation procedure.

To prevent data leakage, feature importance was extracted after model training within each experimental repetition, using only the model fitted to the corresponding augmented training set. When an explanation method required access to evaluation data, it was applied only after model training, and all preprocessing steps were completed. Consequently, any differences in the feature importance rankings can be attributed to the effects of oversampling rather than contamination by information outside the intended data split.

For each model–XAI pair, the pipeline for training, evaluation, and feature importance ranking extraction is independent, enabling parallel execution to reduce the overall runtime. As the computational costs of each predictive model and XAI technique varied significantly, the resulting workload was inherently irregular. To address this, we adopted a task-based parallel programming paradigm, which is well-suited for irregular workloads, in contrast to traditional parallel programming models that assume regular computation patterns. Our implementation is based on torcpy~\cite{torcpy}, an MPI-based Python runtime for task-based parallelism on both shared and distributed memory platforms.

\subsection{Evaluation Metrics} 
\label{sec:evaluation_metrics}

The evaluation considered four complementary dimensions: predictive performance, feature importance preservation, explainability consistency, and class separability. Predictive performance was assessed using Balanced Accuracy and the Matthews Correlation Coefficient (MCC) as the primary threshold-dependent metrics, as both provide robust performance estimates under class imbalance. The F1-score was also used to summarize positive-class performance. To evaluate threshold-independent discrimination, the Precision–Recall Area Under the Curve (PR-AUC) was computed from the predicted probabilities, as it is particularly informative when the minority class is underrepresented. Because the conventional decision threshold of 0.5 is often suboptimal for imbalanced datasets, binary predictions were generated using a threshold that maximized the geometric mean (G-mean) of sensitivity and specificity. This criterion balances minority-class recall and majority-class specificity, providing a more equitable assessment of classification performance.

Feature importance preservation was quantified using Kendall’s $\tau$, which measures the ordinal agreement between the feature importance rankings obtained before and after synthetic oversampling. Finally, explainability consistency was evaluated to determine whether the model-derived feature rankings remained aligned with an external reference ranking. Following the functionality-grounded evaluation framework of Elkhawaga \textit{et al.}~\cite{Elkhawaga2023FeatureEliminationXAI}, reference rankings were constructed from a diverse set of filter-based feature selection methods and used as proxies for ground-truth feature relevance. Model-based and reference rankings were then compared to quantify the extent to which the explanations reflected feature importance supported by the data. The selected filter methods span statistical dependence, information-theoretic, neighborhood-based, impurity-based, and manifold-structure criteria, thereby capturing complementary notions of feature relevance.

\begin{enumerate}
    \item \textbf{Distance Correlation (dCor)}: A nonparametric measure of statistical dependence based on the pairwise distances among observations \cite{Das2024-fy}. Unlike conventional correlation coefficients, which primarily characterize linear relationships, distance correlation can capture both linear and nonlinear associations. For each feature $X_j$, its relevance score was computed as $\operatorname{dCor}(X_j,y)$, where $y$ denotes the binary target variable, and features were ranked in descending order of their distance-correlation values.
    
    \item \textbf{Mutual Information (MI)}: An information-theoretic filter that quantifies the statistical dependency between each feature and the target variable, capturing both linear and non-linear relationships. Features with higher mutual information are considered more informative \cite{Battiti1994MutualInformationFeatureSelection}.
    
    \item \textbf{Laplacian Score}: An unsupervised method that evaluates each feature's ability to preserve local neighborhood structures in the data \cite{He2005LaplacianScore}. It favors features that reflect the intrinsic geometric layout and assigns lower scores to less informative features.
    
    \item \textbf{ReliefF}: A neighbor-based filter that assesses how well features differentiate between nearby instances from different classes \cite{Kononenko1994RELIEF}. For each sample, the algorithm compares the feature values with its nearest neighbors from the same and different classes. It increases a feature’s weight when this feature helps distinguish classes and decreases it when it varies within the same class. ReliefF is robust against noise and sensitive to feature interactions.
    
    \item \textbf{Gini Index}: A measure of impurity used to evaluate the quality of a feature based on how well it separates different classes. It calculates the probability that a random data instance is incorrectly classified when labelled according to the class distribution in a subset. A Gini Index of 0 indicates perfect purity (all data instances belong to one class), while higher values indicate increasing impurity \cite{Breiman1984CART}.
    
    \item \textbf{Joint Mutual Information (JMI)}: An information-theoretic criterion that selects features by maximizing the information shared between a candidate feature, the already selected feature set, and the target class. It is designed to account for feature interactions, aiming to provide a balance between the relevance of a feature to the class and its redundancy with features already in the subset \cite{Yang1999DataVisualizationFeatureSelection}.

    \item \textbf{Maximum Relevance and Max-Independence (MRI)}: A greedy multivariate filter method selects features by considering both their relevance to the target and the amount of new, independent classification information they add concerning the features already selected. MRI emphasizes whether a candidate feature contributes additional class-discriminative information beyond the currently selected subset~\cite{Wang2017MRI}.
\end{enumerate}

\begin{figure}
\centering
    \includegraphics[width=0.7\textwidth]{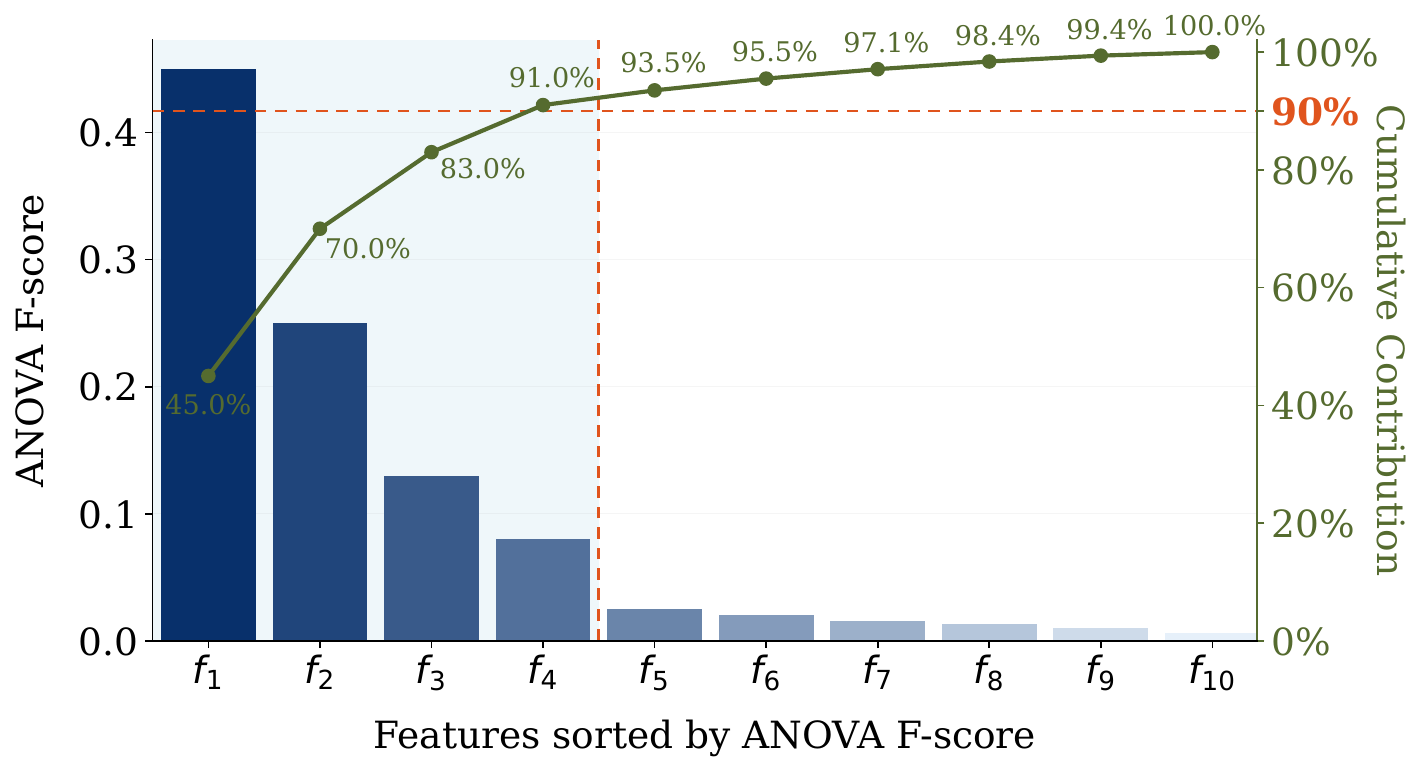}
    \caption{Toy example of feature selection using cumulative ANOVA F-score: the smallest subset reaching 90\% cumulative contribution is selected (here, the first four features).}
    \label{fig:cumulative-anova-feature-selection}
\end{figure}

For each dataset, explainability consistency was measured as the average overlap between the top-ranked features identified by the model and the reference methods. The subset size was determined as the smallest number of top features accounting for at least 90\% of the cumulative ANOVA $F$-score, and this subset was used for the overlap calculation. Fig.~\ref{fig:cumulative-anova-feature-selection} provides a toy example in which, for a dataset with 10 features, the top 4 features account for more than 90\% of the cumulative ANOVA $F$-score and are therefore selected. Table~\ref{tab:selected_subsets} reports the selected subset sizes and their proportions relative to the total number of features for each dataset. Finally, because imbalance severity should not be characterized solely by class ratios \cite{Han2025DataComplexityMeasure}, we also assessed class separability as an additional source of classification difficulty using the $1-N3$ indicator, as described in Section~\ref{datasets}.

\begin{table}
\centering
\caption{Number and proportion of selected features per dataset, defined as the smallest subset accounting for at least 90\% of the cumulative ANOVA F-score.}
\label{tab:selected_subsets}
\begin{tabular}{lccc}
\hline
\textbf{Dataset} & \textbf{\# Selected} & \textbf{Total Features} & \textbf{Proportion} \\
\hline
ai4i 2020 & 3 & 6 & 0.50 \\
abalone & 6 & 8 & 0.75 \\
airlines & 5 & 7 & 0.71 \\
seismic-bumps & 10 & 15 & 0.67 \\
bank customer churn prediction & 4 & 10 & 0.40 \\
churn & 7 & 20 & 0.35 \\
bank-marketing & 5 & 16 & 0.31 \\
rl & 5 & 12 & 0.42 \\
online shoppers purchasing intention & 6 & 17 & 0.35 \\
car\_eval\_4 & 4 & 6 & 0.67 \\
mammography & 3 & 6 & 0.50 \\
ur3 cobotops & 12 & 20 & 0.60 \\
nhanes & 4 & 7 & 0.57 \\
magic gamma telescope & 4 & 10 & 0.40 \\
fried & 3 & 10 & 0.30 \\
bank32nh & 3 & 32 & 0.09 \\
wilt & 3 & 5 & 0.60 \\
lines-overload-50 & 110 & 187 & 0.59 \\
japanese vowels & 6 & 14 & 0.43 \\
pen\_digits & 9 & 16 & 0.56 \\
\hline
\end{tabular}
\end{table}

\section{Experimental Results}
\label{results}

This section evaluates the effectiveness of K-IPO in enhancing minority-class representation while maintaining the original feature importance ranking. The evaluation was conducted in two stages. First, a sensitivity analysis examined how key K-IPO design choices, namely the candidate generator and the strength of the importance preservation constraints, affect the trade-off between predictive performance and interpretability. Second, K-IPO was compared with the baseline oversampling methods using the predictive and interpretability metrics defined above.

\subsection{Sensitivity Analysis of K-IPO Design Components}
\label{sens_analysis}

We first examined the sensitivity of K-IPO to its two main design components: the candidate generator and importance preservation constraints. At the beginning of each repetition, the original data were stratified into training and test sets using an 80/20 ratio. All preprocessing transformations were fitted exclusively to the training set and subsequently applied to the test set. Each of the three classifiers was then fitted independently to the resulting augmented training set. The classification threshold that maximized the G-mean was estimated from an internal validation split of the training data and fixed before evaluation. Finally, the predictive performance and model-derived feature importance were evaluated using the untouched original test set.

\begin{sidewaystable}[p]
  \centering
  \caption{Generator-sensitivity analysis of K-IPO at a fixed Kendall's agreement threshold of $\tau_0 = 0.7$. Reported values are averaged across 10 independent repetitions and three classifiers.}
  \label{tab:kipo-generators}

  \scriptsize
  \setlength{\tabcolsep}{2.2pt}
  \renewcommand{\arraystretch}{0.90}

  \resizebox{\linewidth}{!}{%
  \begin{tabular}{l*{4}{ccccc}}
    \toprule
    \multirow{2}{*}{\textbf{Dataset}} &
    \multicolumn{5}{c}{\textbf{Balanced Accuracy}} &
    \multicolumn{5}{c}{\textbf{F1-score}} &
    \multicolumn{5}{c}{\textbf{PR AUC}} &
    \multicolumn{5}{c}{\textbf{MCC}} \\
    \cmidrule(lr){2-6}
    \cmidrule(lr){7-11}
    \cmidrule(lr){12-16}
    \cmidrule(lr){17-21}
    & SMOTE & CTGAN & TVAE & GC & TabDDPM &
      SMOTE & CTGAN & TVAE & GC & TabDDPM &
      SMOTE & CTGAN & TVAE & GC & TabDDPM &
      SMOTE & CTGAN & TVAE & GC & TabDDPM \\
    \midrule

    \multicolumn{21}{l}{\textbf{Mixed numerical--categorical datasets}} \\
    \midrule

    D1  & \textbf{0.9774} & \underline{0.9659} & 0.9626 & 0.9655 & 0.8996 &
           \textbf{0.9730} & 0.9616 & 0.9558 & \underline{0.9618} & 0.8865 &
           \underline{0.9916} & \textbf{0.9923} & 0.9897 & 0.9913 & 0.9540 &
           \textbf{0.9526} & 0.9332 & 0.9221 & \underline{0.9338} & 0.8057 \\

    D2  & 0.8954 & \textbf{0.9311} & \underline{0.9056} & 0.8436 & $-$ &
           0.8802 & \textbf{0.9241} & \underline{0.8922} & 0.8238 & $-$ &
           0.8967 & \textbf{0.9787} & \underline{0.9640} & 0.8617 & $-$ &
           0.7847 & \textbf{0.8740} & \underline{0.8121} & 0.6803 & $-$ \\

    D3  & \textbf{0.8263} & 0.7847 & \underline{0.8210} & 0.7769 & $-$ &
           \textbf{0.8033} & 0.7557 & \underline{0.7954} & 0.7406 & $-$ &
           \underline{0.8566} & 0.8380 & \textbf{0.8762} & 0.8168 & $-$ &
           \textbf{0.6534} & 0.5691 & \underline{0.6447} & 0.5555 & $-$ \\

    D4  & \underline{0.9016} & \textbf{0.9518} & $-$ & $-$ & $-$ &
           \underline{0.8874} & \textbf{0.9482} & $-$ & $-$ & $-$ &
           \underline{0.9317} & \textbf{0.9769} & $-$ & $-$ & $-$ &
           \underline{0.7997} & \textbf{0.9137} & $-$ & $-$ & $-$ \\

    D5  & \underline{0.9036} & \textbf{0.9100} & $-$ & $-$ & $-$ &
           \underline{0.8893} & \textbf{0.8987} & $-$ & $-$ & $-$ &
           \underline{0.9507} & \textbf{0.9581} & $-$ & $-$ & $-$ &
           \underline{0.8033} & \textbf{0.8290} & $-$ & $-$ & $-$ \\

    D6  & \textbf{0.9520} & $-$ & $-$ & \underline{0.8874} & $-$ &
           \textbf{0.9446} & $-$ & $-$ & \underline{0.8695} & $-$ &
           \textbf{0.9850} & $-$ & $-$ & \underline{0.9390} & $-$ &
           \textbf{0.9030} & $-$ & $-$ & \underline{0.7801} & $-$ \\

    D7  & 0.9037 & \textbf{0.9120} & $-$ & \underline{0.9073} & $-$ &
           0.8898 & \textbf{0.9007} & $-$ & \underline{0.8957} & $-$ &
           0.9492 & \textbf{0.9675} & $-$ & \underline{0.9643} & $-$ &
           0.8052 & \textbf{0.8304} & $-$ & \underline{0.8219} & $-$ \\

    D8  & \textbf{0.9032} & \underline{0.8978} & $-$ & 0.8558 & $-$ &
           \textbf{0.8901} & \underline{0.8842} & $-$ & 0.8354 & $-$ &
           \underline{0.9434} & \textbf{0.9452} & $-$ & 0.9183 & $-$ &
           \textbf{0.8094} & \underline{0.8063} & $-$ & 0.7124 & $-$ \\

    D9  & \underline{0.9362} & \textbf{0.9447} & $-$ & $-$ & $-$ &
           \underline{0.9256} & \textbf{0.9383} & $-$ & $-$ & $-$ &
           \underline{0.9690} & \textbf{0.9847} & $-$ & $-$ & $-$ &
           \underline{0.8682} & \textbf{0.8941} & $-$ & $-$ & $-$ \\

    \midrule
    \multicolumn{21}{l}{\textbf{Categorical-only datasets}} \\
    \midrule

    D10 & \textbf{1.0000} & 0.9850 & \underline{0.9995} & 0.9850 & 0.4742 &
           \textbf{1.0000} & 0.9804 & \underline{0.9994} & 0.9804 & 0.4310 &
           \textbf{1.0000} & 0.9849 & \underline{0.9999} & 0.9856 & 0.4456 &
           \textbf{1.0000} & 0.9657 & \underline{0.9989} & 0.9657 & -0.0512 \\

    \midrule
    \multicolumn{21}{l}{\textbf{Numerical-only datasets}} \\
    \midrule

    D11 & \textbf{0.9871} & 0.9786 & \underline{0.9852} & $-$ & 0.9361 &
           \textbf{0.9839} & 0.9753 & \underline{0.9821} & $-$ & 0.9281 &
           0.9949 & \textbf{0.9966} & \underline{0.9954} & $-$ & 0.9692 &
           \textbf{0.9718} & 0.9567 & \underline{0.9685} & $-$ & 0.8763 \\

    D12 & \underline{0.9436} & $-$ & $-$ & \textbf{0.9566} & $-$ &
           \underline{0.9345} & $-$ & $-$ & \textbf{0.9521} & $-$ &
           \underline{0.9736} & $-$ & $-$ & \textbf{0.9877} & $-$ &
           \underline{0.8838} & $-$ & $-$ & \textbf{0.9177} & $-$ \\

    D13 & 0.8774 & \underline{0.8944} & $-$ & \textbf{0.9304} & 0.7237 &
           0.8612 & \underline{0.8813} & $-$ & \textbf{0.9232} & 0.6861 &
           \underline{0.9530} & 0.9369 & $-$ & \textbf{0.9663} & 0.7618 &
           0.7578 & \underline{0.8008} & $-$ & \textbf{0.8726} & 0.4480 \\

    D14 & \textbf{0.9772} & \underline{0.9656} & 0.9387 & 0.9507 & 0.8978 &
           \textbf{0.9741} & \underline{0.9619} & 0.9299 & 0.9442 & 0.8836 &
           \textbf{0.9961} & \underline{0.9900} & 0.9783 & 0.9850 & 0.9561 &
           \textbf{0.9547} & \underline{0.9341} & 0.8779 & 0.9029 & 0.7990 \\

    D15 & \textbf{0.9696} & 0.9430 & \underline{0.9595} & 0.8676 & 0.8849 &
           \textbf{0.9651} & 0.9366 & \underline{0.9547} & 0.8484 & 0.8689 &
           \textbf{0.9946} & 0.9803 & \underline{0.9895} & 0.9333 & 0.9483 &
           \textbf{0.9389} & 0.8919 & \underline{0.9215} & 0.7454 & 0.7785 \\

    D16 & \textbf{0.9758} & $-$ & $-$ & $-$ & $-$ &
           \textbf{0.9720} & $-$ & $-$ & $-$ & $-$ &
           \textbf{0.9828} & $-$ & $-$ & $-$ & $-$ &
           \textbf{0.9512} & $-$ & $-$ & $-$ & $-$ \\

    D17 & \textbf{0.9910} & 0.9436 & 0.9387 & 0.9648 & \underline{0.9768} &
           \textbf{0.9890} & 0.9357 & 0.9305 & 0.9592 & \underline{0.9722} &
           \textbf{0.9980} & 0.9819 & 0.9797 & 0.9907 & \underline{0.9943} &
           \textbf{0.9807} & 0.8881 & 0.8794 & 0.9287 & \underline{0.9512} \\

    D18 & \textbf{0.9979} & 0.9955 & \underline{0.9969} & 0.9964 & $-$ &
           \textbf{0.9971} & 0.9944 & \underline{0.9962} & 0.9954 & $-$ &
           \underline{0.9990} & \textbf{0.9998} & \textbf{0.9998} & \textbf{0.9998} & $-$ &
           \textbf{0.9950} & 0.9901 & \underline{0.9933} & 0.9919 & $-$ \\

    D19 & \textbf{0.9977} & 0.9887 & \underline{0.9891} & 0.9858 & 0.9876 &
           \textbf{0.9973} & \underline{0.9871} & \underline{0.9871} & 0.9838 & 0.9858 &
           \textbf{0.9999} & \underline{0.9990} & \underline{0.9990} & 0.9985 & 0.9988 &
           \textbf{0.9953} & 0.9774 & \underline{0.9775} & 0.9716 & 0.9752 \\

    D20 & \textbf{0.99998} & 0.9975 & \underline{0.9981} & $-$ & 0.9909 &
           \textbf{0.99998} & 0.9972 & \underline{0.9978} & $-$ & 0.9898 &
           \textbf{1.0000} & 0.9997 & \underline{0.9999} & $-$ & 0.9989 &
           \textbf{0.99996} & 0.9951 & \underline{0.9962} & $-$ & 0.9821 \\

    \midrule

    \multicolumn{21}{l}{%
      \textbf{Wins:}
      SMOTE = (\textbf{13}, \textbf{13}, \underline{8}, \textbf{13}),\;
      CTGAN = (\underline{5}, \underline{5}, \textbf{9}, \underline{5}),\;
      TVAE = (0, 0, 2, 0),\;
      GC = (2, 2, 3, 2),\;
      TabDDPM = (0, 0, 0, 0)
      \quad
      \textbf{Failures:}
      SMOTE = 0,\; CTGAN = 3,\; TVAE = 9,\; GC = 6,\; TabDDPM = 11
    } \\

    \bottomrule
  \end{tabular}%
  }

  \vspace{1mm}
  \begin{minipage}{\linewidth}
  \scriptsize
  \textit{Note.} Bold and underlined values indicate the best- and second-best-performing generators, respectively, for each dataset and metric. A dash ($-$) indicates failure to generate the requested $N^\star$ accepted synthetic samples within a practical execution time. Failures count such non-completions across datasets. Wins are ordered as Balanced Accuracy, F1-score, PR AUC, and MCC. The SMOTE-family variant was selected according to feature composition: SMOTENC for mixed numerical--categorical datasets, SMOTEN for categorical-only datasets, and SMOTEWB for numerical-only datasets.
  \end{minipage}
\end{sidewaystable}

Regarding the most suitable candidate data generator within K-IPO, we compared the methods discussed in Section~\ref{baseline_methods}. Performance was assessed using Balanced Accuracy, F1-score, PR-AUC, and MCC. Table~\ref{tab:kipo-generators} reports the corresponding results, with the best and second-best values for each dataset--metric pair shown in bold and underlined, respectively. In this experiment, Kendall’s agreement threshold was fixed at $\tau_0 = 0.7$, enforcing a strong but not overly restrictive agreement between the original and augmented feature importance rankings. This configuration allows K-IPO to reject synthetic samples that distort the feature importance structure while still permitting sufficient flexibility for minority-class augmentation. The reported values were averaged across repetitions and the three classifiers. A failure was recorded whenever K-IPO could not obtain the target number $N^\star$ of accepted synthetic samples within a reasonable runtime under the available computational resources. These cases are indicated by a dash in Table~\ref{tab:kipo-generators}.

The results indicate that K-IPO performs more consistently when paired with SMOTE-based generators. Specifically, SMOTE achieved the highest number of wins for Balanced Accuracy, F1-score, and MCC, with 13 wins in each category, while also securing the second-highest number of PR-AUC wins (8). Notably, SMOTE was the only generator that completed the requested augmentation for all 20 datasets, indicating reliable candidate generation under the experimental conditions. This can be attributed to the generation of synthetic samples in the local neighborhood of existing minority class observations, thereby preserving the minority-class geometry and producing candidates that are more likely to satisfy the imposed constraints. This behavior is particularly advantageous in small- and medium-sized tabular datasets, where more complex deep generative models may not have sufficient minority class evidence to learn stable data distributions \cite{Wang2024GenerativeModelsTabularData, Won2026SyntheticDataAugmentation}.

Among the deep generative models, CTGAN was the strongest performer. It achieved the second-highest number of wins in terms of Balanced Accuracy, F1-score, and MCC (5 wins each) and the highest number of wins in terms of PR-AUC (9 wins). This indicates that CTGAN is particularly effective when performance depends on the quality of probability estimates and instance ranking, likely because its conditional generation mechanism can capture complex nonlinear dependencies in tabular data. However, CTGAN also recorded 3 failures and exhibited lower overall robustness, suggesting that its greater expressive power comes at the cost of increased training instability \cite{HayaeianShirvan2025DeepGenerativeOversampling}. Gaussian Copula was less competitive, achieving only 2 wins in Balanced Accuracy, F1-score, and MCC, and 3 wins in PR-AUC, while failing on 6 datasets. These results highlight the limitations of its parametric dependence assumptions, which may be insufficient for heterogeneous datasets with complex nonlinear relationships. TVAE showed limited effectiveness, with no wins in Balanced Accuracy, F1-score, or MCC, only 2 wins in PR-AUC, and 9 failures. Although VAEs are generally considered more stable and easier to train than GANs, latent space regularization often produces overly smooth synthetic samples, reducing their discriminative value \cite{Takida2022VAEGeneralizedVariance}. This drawback becomes particularly problematic in highly imbalanced settings, where VAEs frequently underperform both GAN-based approaches and simpler statistical generators \cite{Kiran2023GANVAEGenerators}. TabDDPM was the weakest method overall, achieving no wins on any metric and failing on 11 datasets, indicating its poor compatibility with the proposed framework. Overall, the results suggest that within K-IPO, simpler interpolation-based generators provide the best trade-off between predictive performance, stability, and compatibility with importance-preserving acceptance. More expressive deep generative models can deliver occasional performance gains, but their benefits are offset by reduced robustness and greater sensitivity across diverse and imbalanced tabular datasets.

Subsequently, we examined the importance of preservation constraints, focusing on the Kendall's threshold $\tau_{0}$ and the top-$k$ ordering constraint. We emphasized top-$k$ ordering because it provides the most informative middle ground between the two alternative top-$k$ constraints. On the one hand, top-$k$ overlap is comparatively weak because it requires only that the same features appear in the top set without preserving their internal order. Moreover, part of this information is already reflected by Kendall’s $\tau$, as a high global rank correlation generally implies substantial agreement among the most influential features. Additionally, top-$k$ overlap does not scale monotonically in an interpretable way with $k$: preserving the top-$k$ set for a larger value of $k$ does not necessarily guarantee the preservation of smaller, more critical subsets, limiting its interpretability as a progressive constraint. In contrast, top-$k$ magnitude preservation is considerably stricter because it constrains relative importance differences in addition to the rank order. This increases the computational burden, especially for larger and high-dimensional datasets that require repeated candidate evaluations, while offering limited additional benefit to the objective of this study, where preserving the order of influential features is the primary concern.

\begin{figure}[t]
  \centering
  \begin{adjustbox}{max width=\textwidth, max totalheight=0.86\textheight, center}
    \begin{minipage}{\textwidth}
      \centering
        \captionsetup[subfigure]{font=footnotesize, labelfont=bf, skip=2pt}
       
        \heatmap{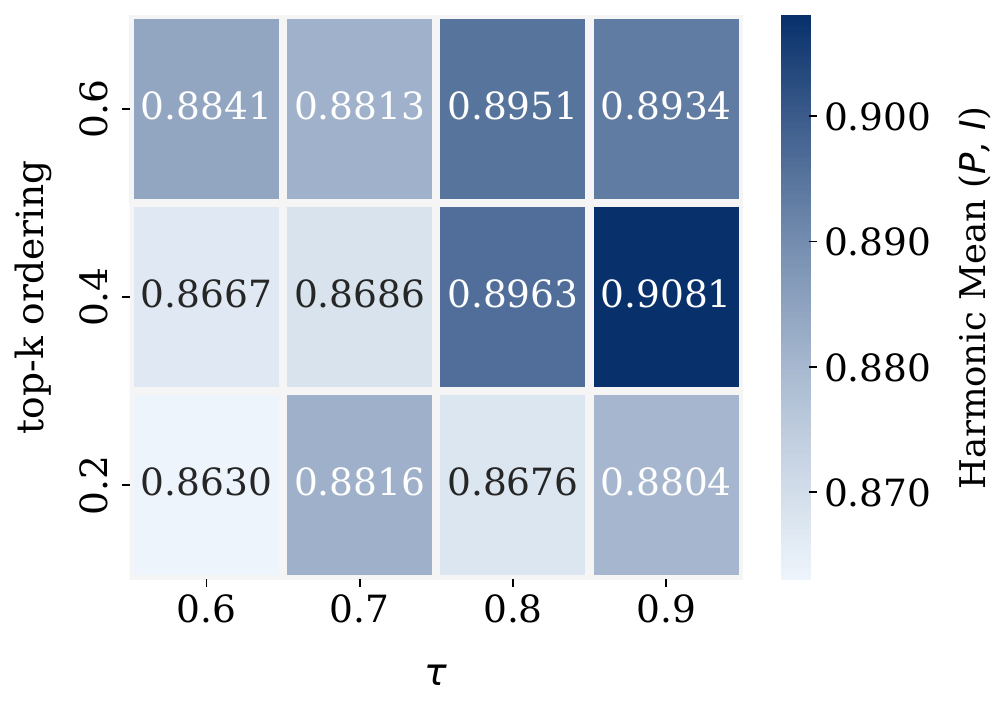}{abalone}\hfill
        \heatmap{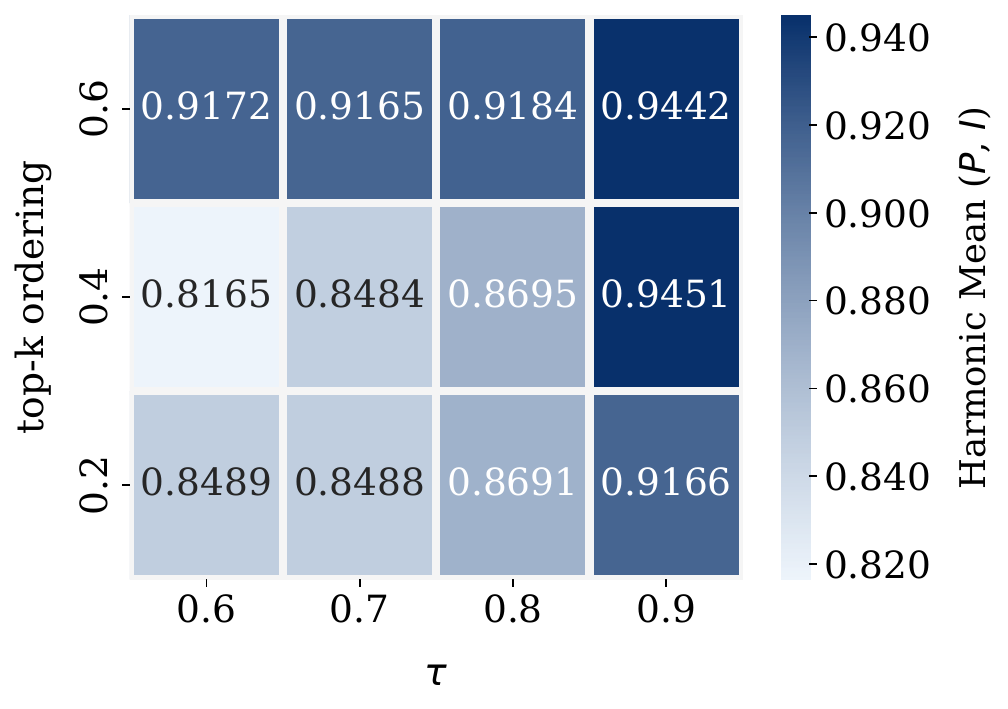}{ai4i 2020}\hfill
        \heatmap{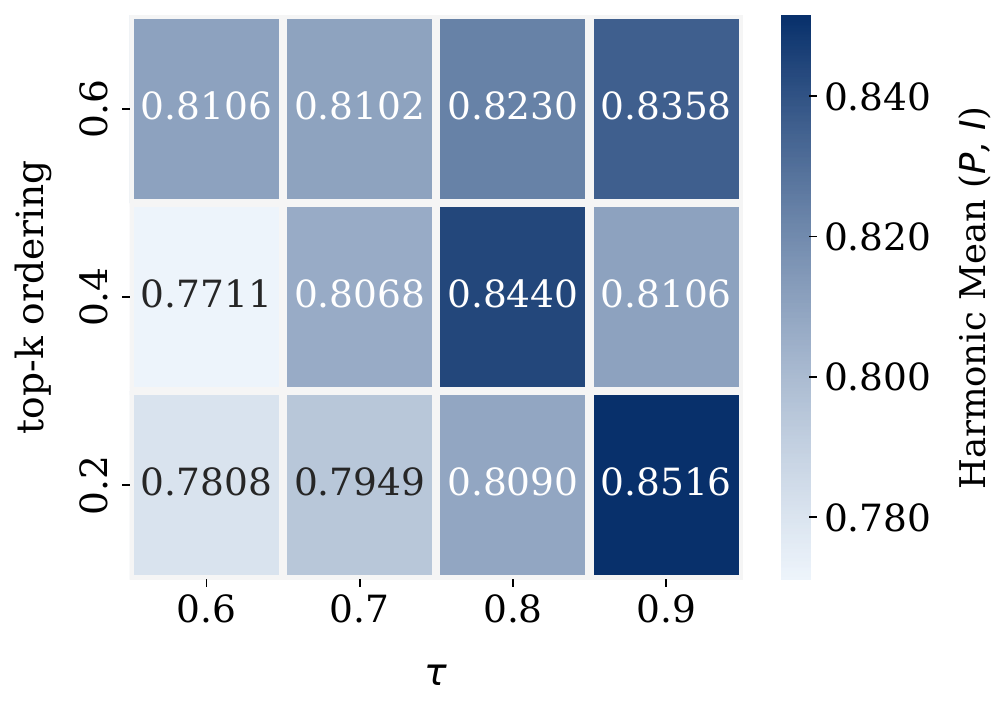}{airlines}\hfill
        \heatmap{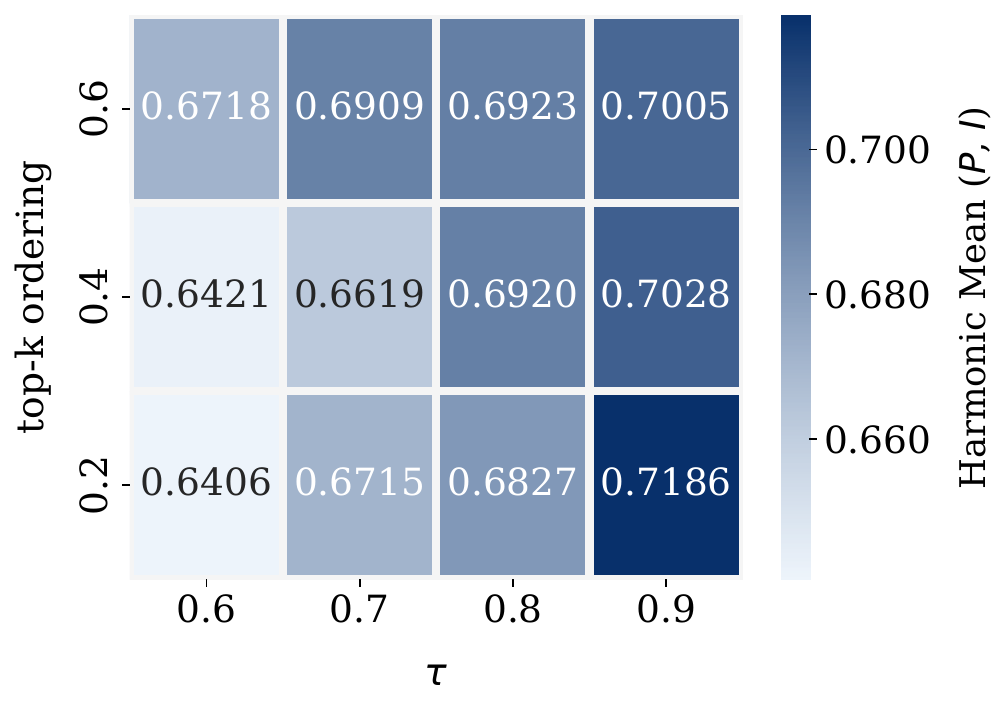}{bank customer churn prediction}\hfill
        
        \vspace{0.3em}
        
        \heatmap{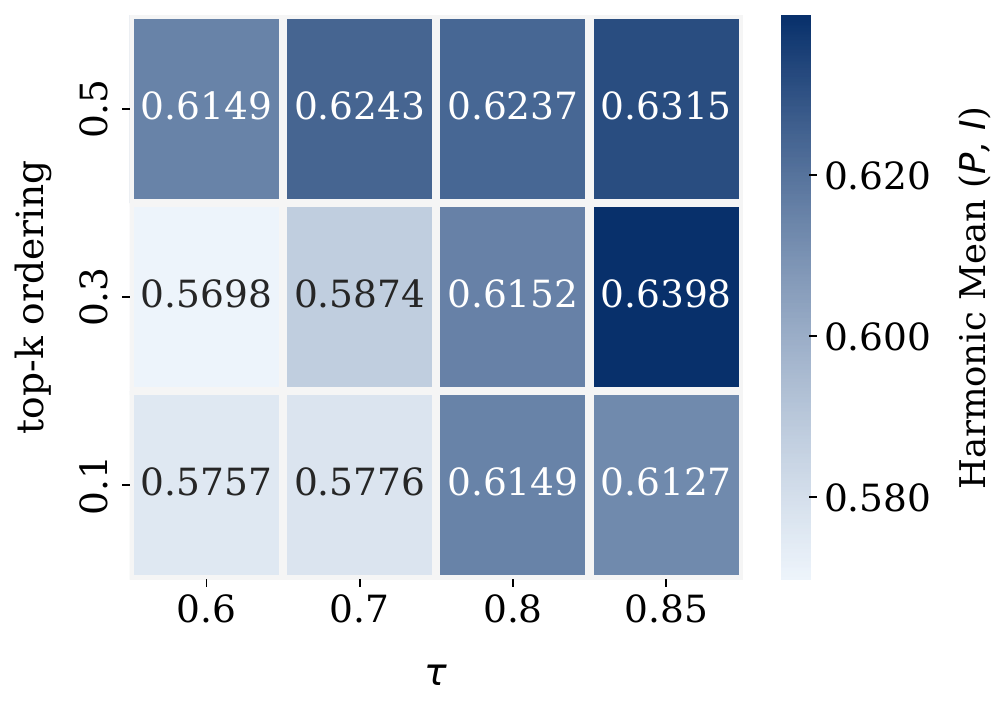}{bank-marketing}\hfill
        \heatmap{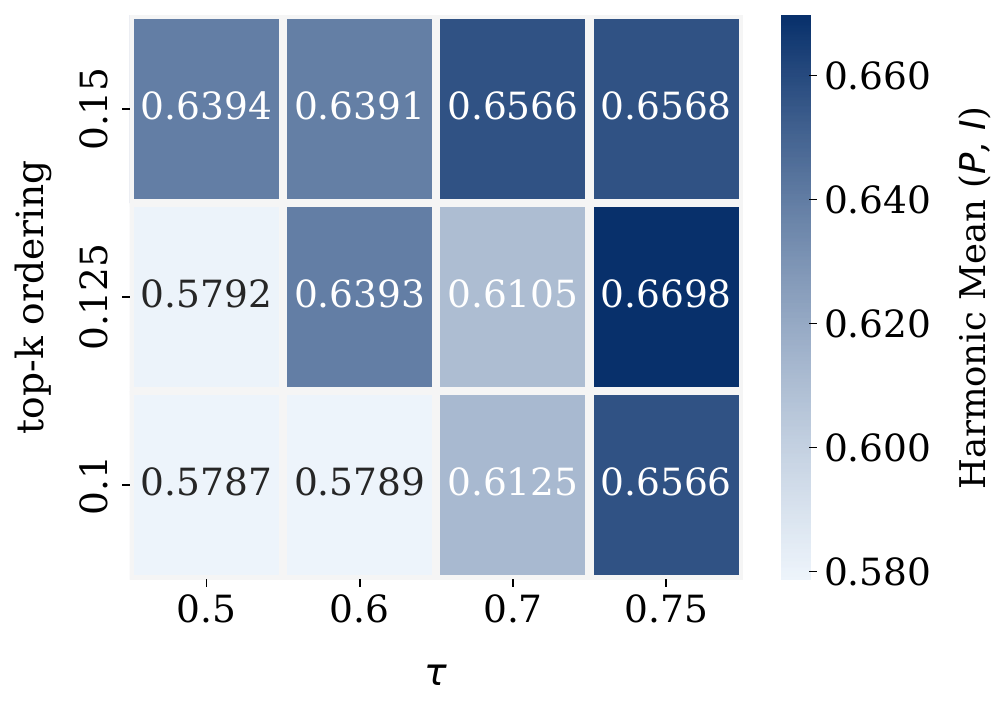}{bank32nh}\hfill
        \heatmap{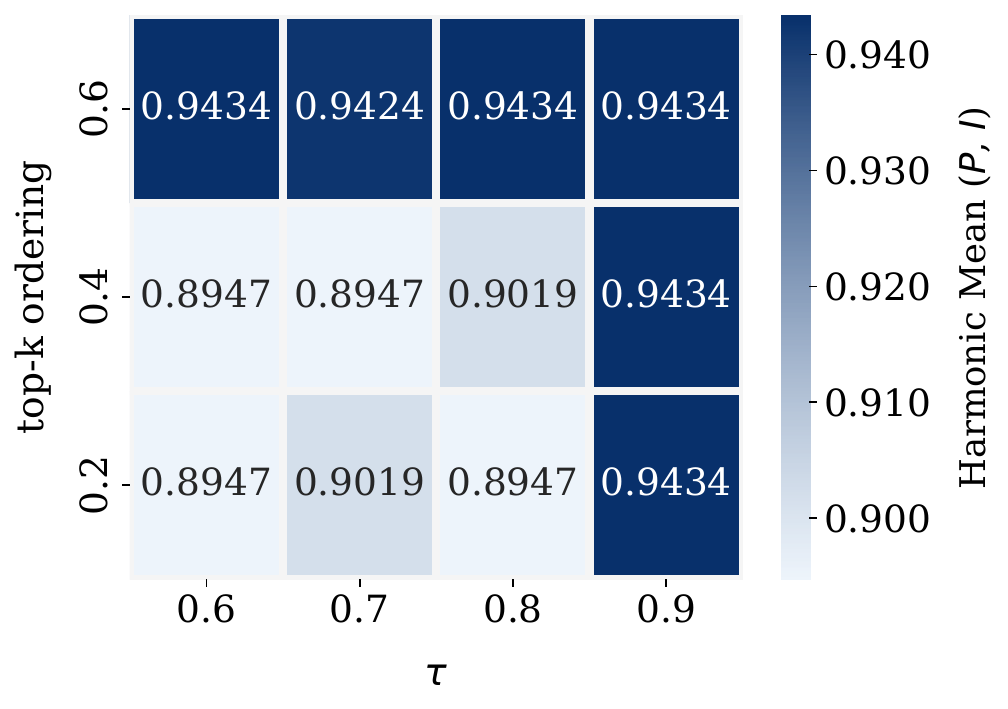}{car\_eval\_4}\hfill
        \heatmap{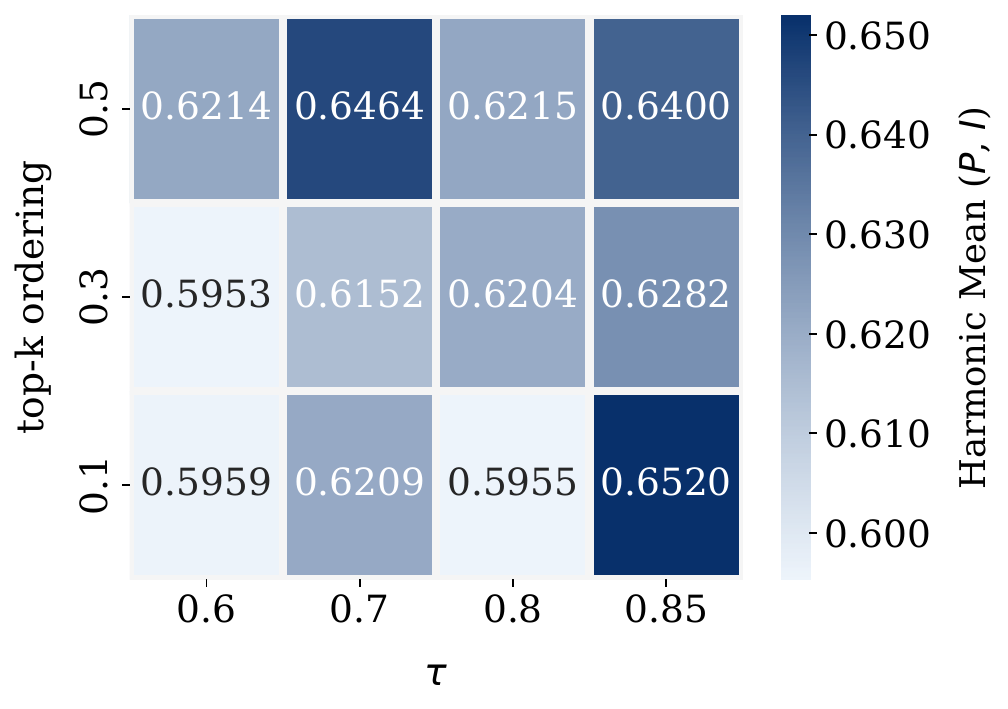}{churn}\hfill
        
          \vspace{0.3em}
        
        \heatmap{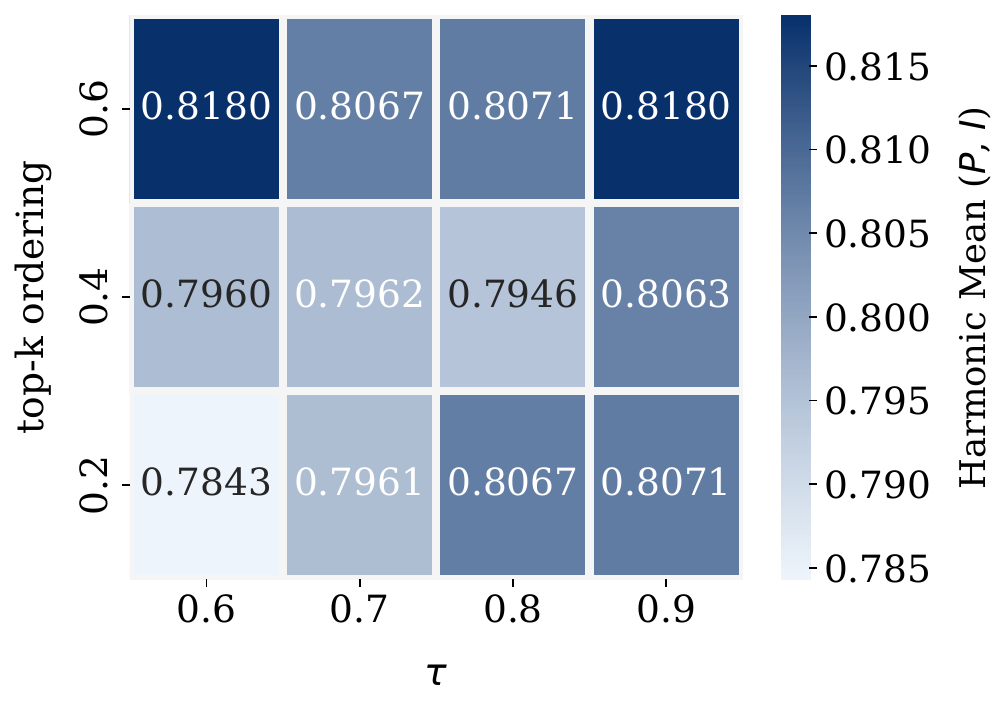}{fried}\hfill
        \heatmap{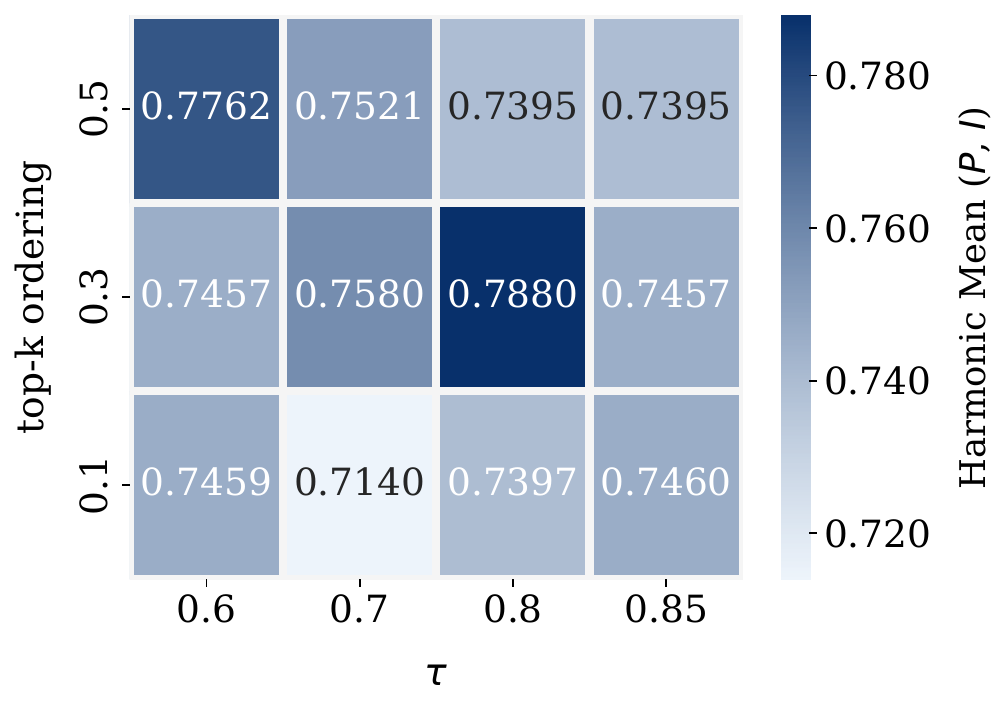}{japanese vowels}\hfill
        \heatmap{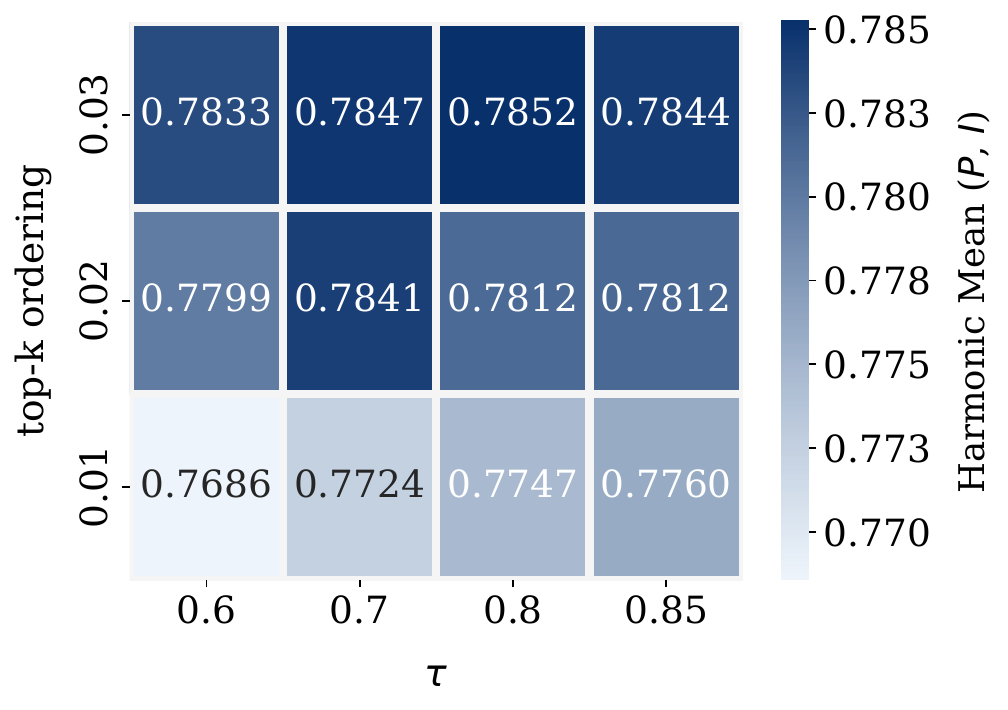}{lines-overload-50}\hfill
        \heatmap{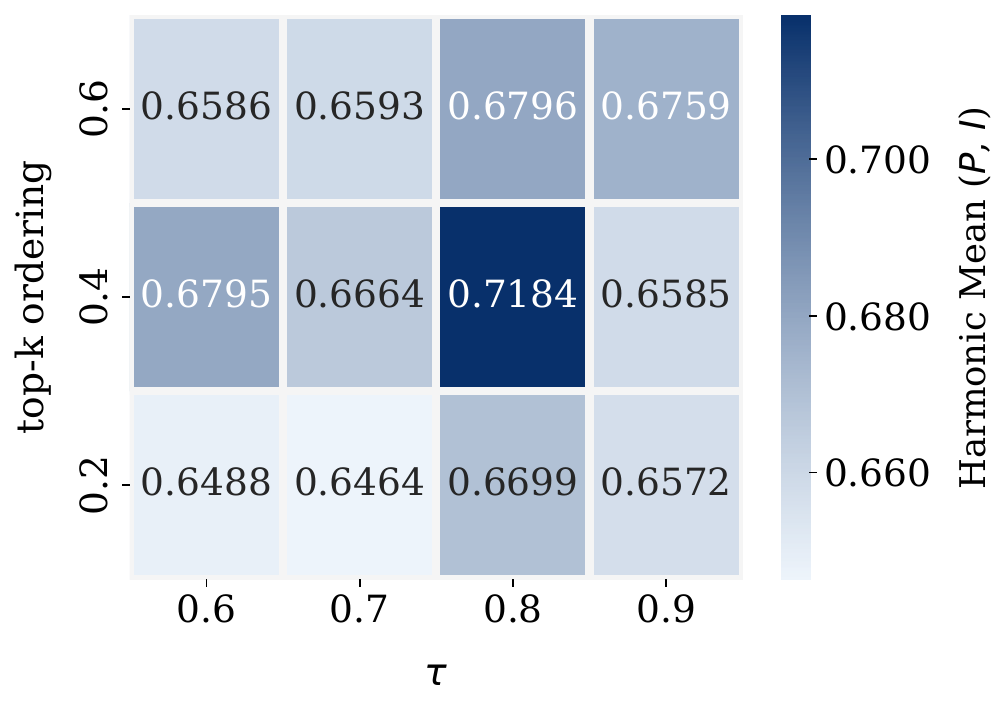}{magic gamma telescope}

        \vspace{0.3em}
        
        \heatmap{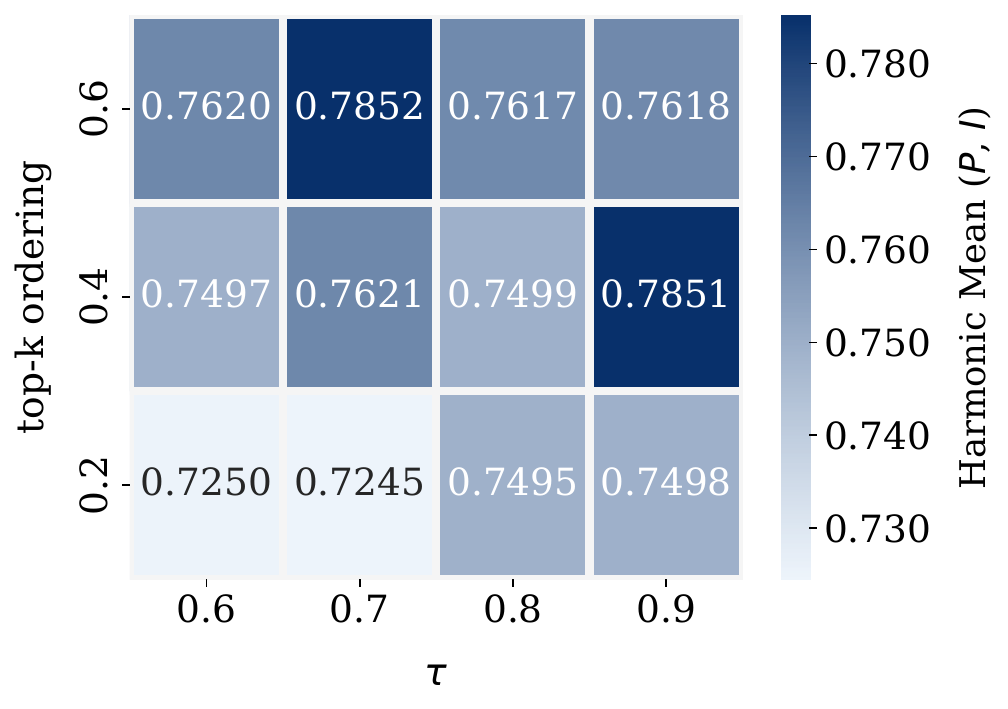}{mammography}\hfill
        \heatmap{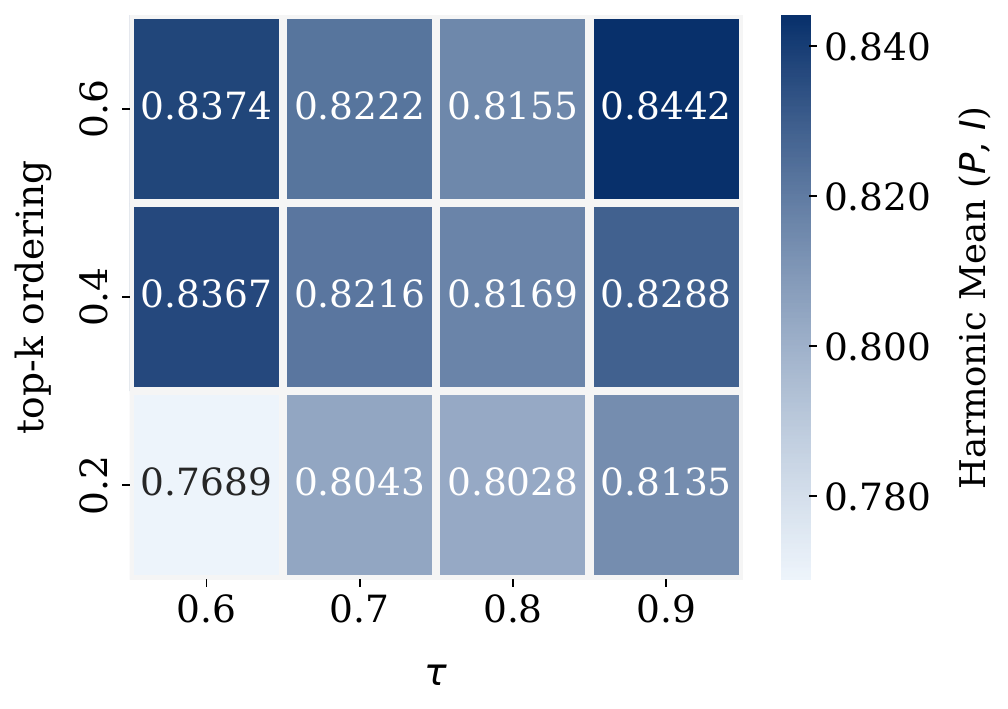}{nhanes}\hfill
        \heatmap{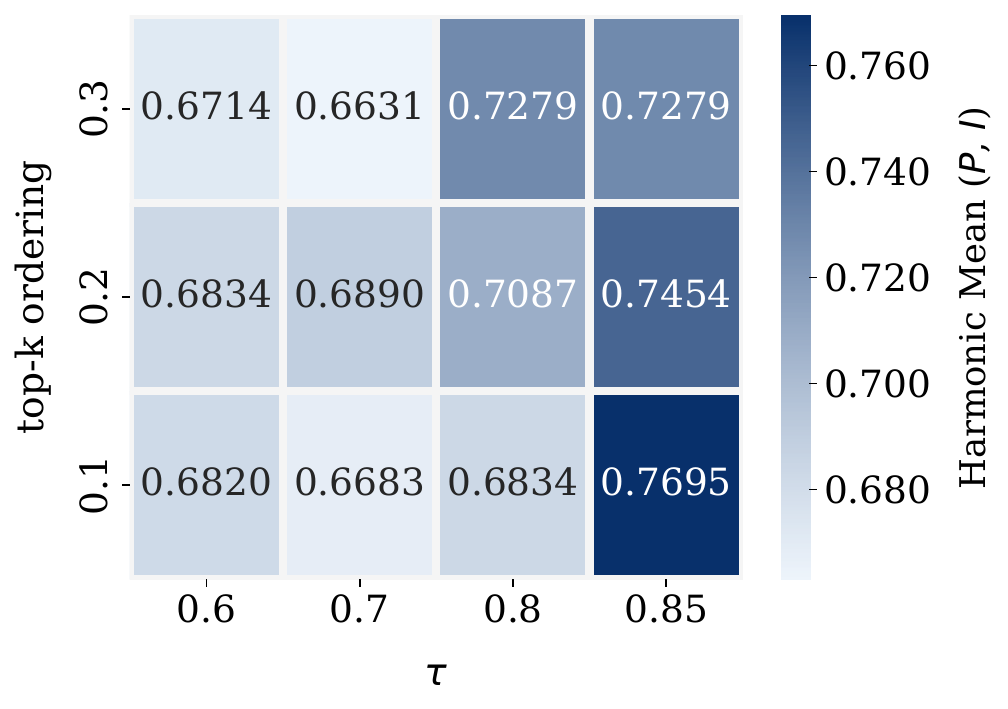}{online shoppers purchasing intention}\hfill
        \heatmap{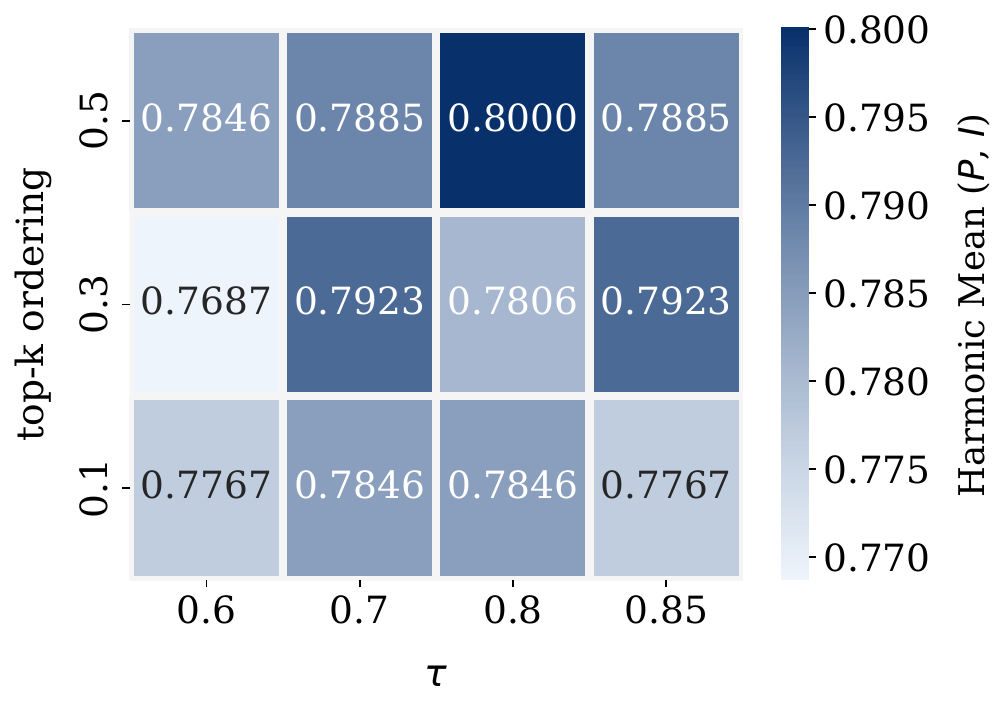}{pen\_digits}

        \vspace{0.3em}
        
        \heatmap{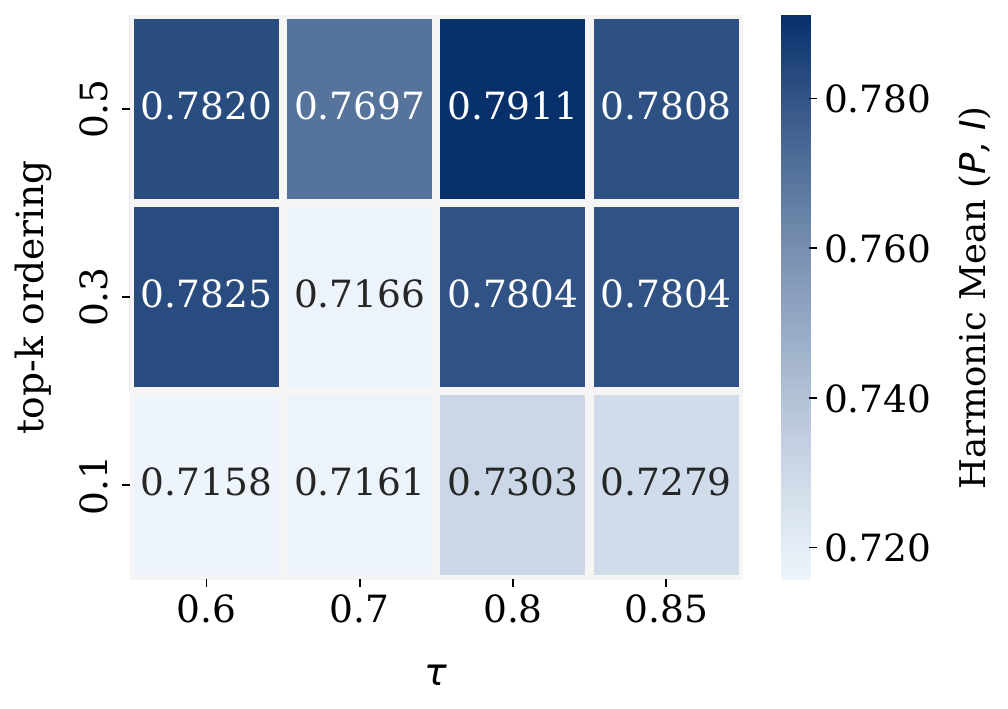}{rl}
        \heatmap{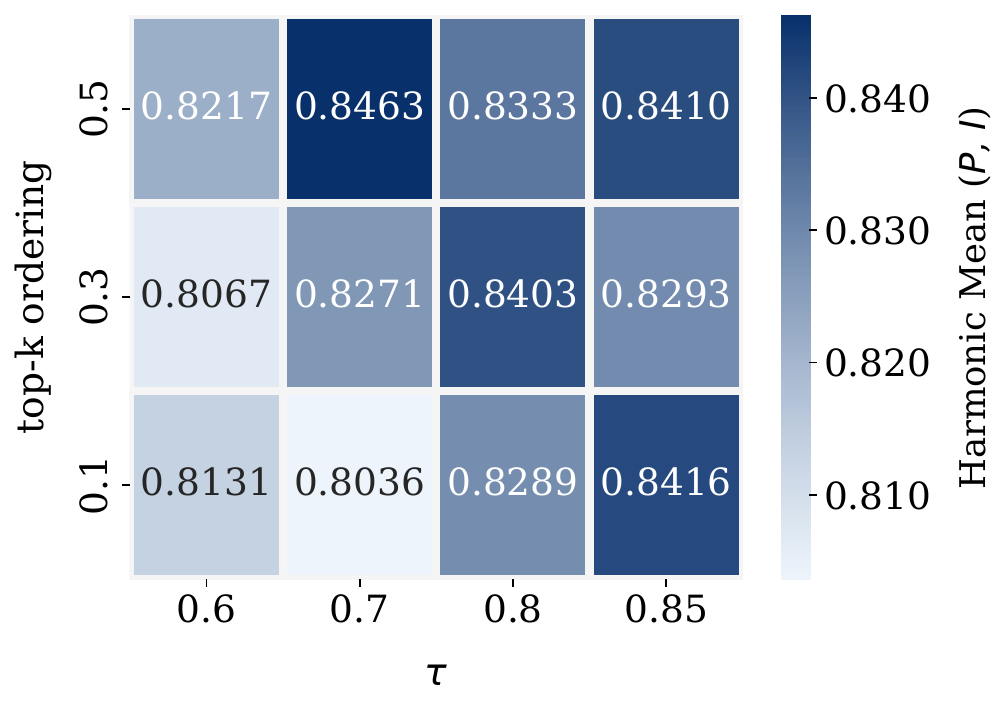}{seismic-bumps}\hfill
        \heatmap{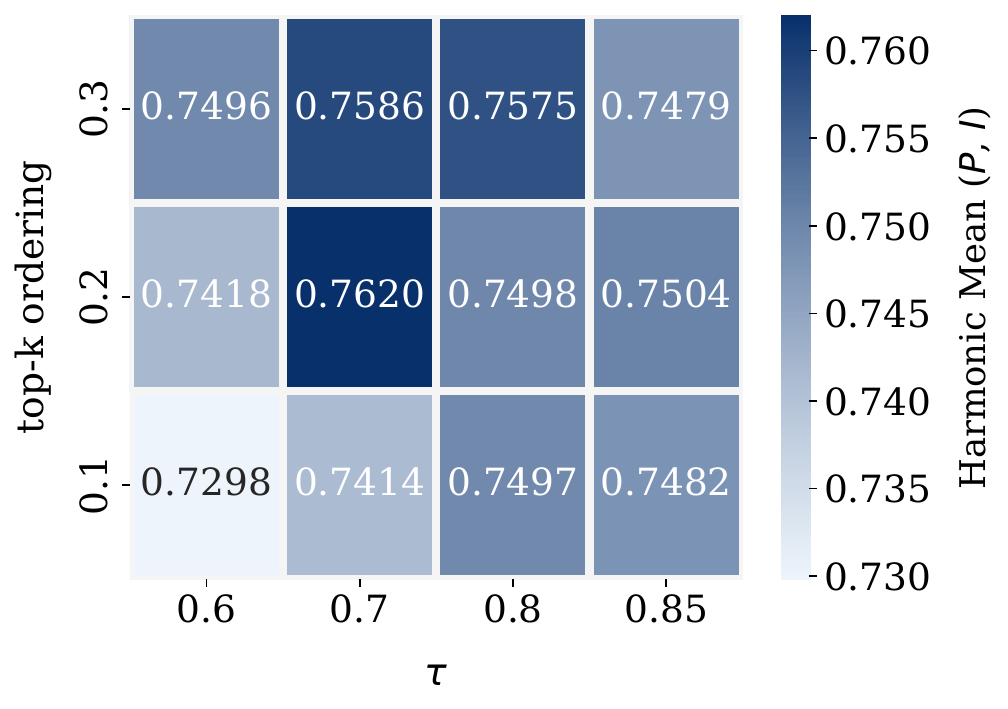}{ur3 cobotops}\hfill
        \heatmap{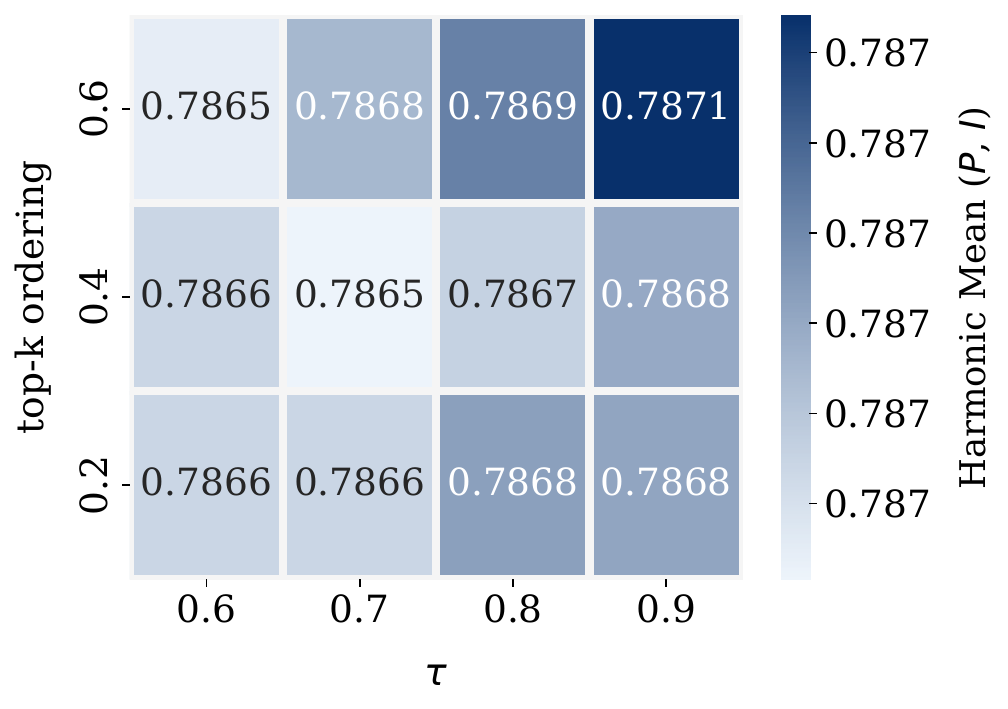}{wilt}
    \end{minipage}
  \end{adjustbox}
  \caption{%
  Harmonic mean $H(P,I)$ heatmaps over $(\tau, \text{top-}k\text{ ordering})$
  for the benchmark datasets. The decimal values in the top-$k$ ordering axis denote the proportion of top-ranked features whose ordering is constrained, e.g., $0.2$ corresponds to the top $20\%$ of features.
  }
  \label{fig:hm:part1}
\end{figure}

To jointly assess predictive performance and interpretability, we computed the harmonic mean of a predictive component $P$ and an interpretability component $I$ for each configuration of $\tau_{0}$ and top-$k$ ordering across all datasets. The predictive component was defined as the average of Accuracy and PR-AUC, while the interpretability component measured the consistency between model-derived and reference feature rankings, as described in Section~\ref{sec:evaluation_metrics}. We combined these components as $\frac{2PI}{P+I}$, favoring configurations that achieved strong performance in both dimensions rather than improving one at the expense of the other.

The results, summarized in Fig.~\ref{fig:hm:part1}, reveal a consistent overall trend, along with dataset-specific variations. Generally, the harmonic mean improves as Kendall’s threshold becomes stricter, indicating that stronger preservation of the original importance ranking enhances the aggregate predictive and explanatory power of the trained models. In contrast, the effect of the top-$k$ ordering constraint is weaker and more dependent on the dataset. In several cases, including airlines, pen\_digits, and seismic-bumps, performance improves as the top-$k$ ratio increases but eventually plateaus, even exhibiting diminishing returns. Across all datasets, increasing the top-$k$ ratio generally leads to performance improvements or stabilization; however, the rate of improvement and saturation point vary, reflecting differences in dataset sensitivity to this constraint. Overall, these findings indicate that the most substantial gains arise from enforcing a sufficiently strict, global rank-preservation constraint. While preserving the ordering of a larger subset of important features is beneficial, the advantages of expanding the top-$k$ constraint tend to saturate, with limited additional gains beyond a dataset-dependent threshold.

\subsection{Comparison with baseline oversampling techniques}

After analyzing the sensitivity of K-IPO to its design parameters, we compared it with the baseline methods described in Section~\ref{baseline_methods}. All methods were applied before model training. For K-IPO, the importance-preserving parameter values were selected separately for each dataset based on the aforementioned sensitivity analysis, ensuring operation within the most suitable regime. In contrast, all baseline methods were evaluated using their default library-recommended configuration parameters to maintain a fair and method-agnostic comparison. As in Section~\ref{sens_analysis}, the augmentation process for each dataset–generator pair was repeated 10 times, and the results are reported as the average over all runs.

Tables \ref{tab:results_d1_d10} and \ref{tab:results_d11_d20} summarize the results across all datasets, reporting the Balanced Accuracy, F1-score, MCC, composite predictive score ($P$), explainability consistency score ($I$), Kendall’s $\tau$ after augmentation, and separability indicator ($1-N3$). All values represent the averages of the three predictive models coupled with their corresponding XAI methods for 10 independent experimental repetitions.

\begin{sidewaystable}[p]
\caption{%
Comparison of synthetic data generation methods on datasets D1--D10 across seven evaluation metrics.
Entries represent mean \texorpdfstring{$\pm$}{±} standard deviation over multiple runs.
\textbf{Bold} values indicate the best performance per metric per dataset.
}
\label{tab:results_d1_d10}
\centering
\scriptsize
\setlength{\tabcolsep}{3pt}
\renewcommand{\arraystretch}{1.1}

\resizebox{\linewidth}{!}{%
\begin{tabular}{l||ccccccc||ccccccc}
\toprule
& \multicolumn{7}{c||}{\textbf{Datasets D1--D5}}
& \multicolumn{7}{c}{\textbf{Datasets D6--D10}} \\
\cmidrule(lr){2-8}\cmidrule(lr){9-15}
\textbf{Method}
& \textbf{Bal.\ Acc.} & \textbf{F1} & \textbf{MCC} & $\boldsymbol{P}$ & $\boldsymbol{I}$ & \textbf{Kendall's} $\boldsymbol{\tau}$ & $\boldsymbol{1\!-\!N3}$
& \textbf{Bal.\ Acc.} & \textbf{F1} & \textbf{MCC} & $\boldsymbol{P}$ & $\boldsymbol{I}$ & \textbf{Kendall's} $\boldsymbol{\tau}$ & $\boldsymbol{1\!-\!N3}$ \\
\midrule

& \multicolumn{7}{l||}{\textit{D1}} & \multicolumn{7}{l}{\textit{D6}} \\
\quad CTGAN
& \rs{0.9567}{0.0097} & \rs{0.9512}{0.0113} & \rs{0.9152}{0.0197} & \rs{0.9738}{0.0068} & \rs{0.6318}{0.0891} & \rs{0.2267}{0.2335} & \rs{0.9302}{ 0.0138}
& \rsb{0.9673}{0.0018} & \rsb{0.9634}{0.0020} & \rsb{0.9366}{0.0035} & \rsb{0.9790}{0.0009} & \rs{0.4082}{0.0314} & \rs{0.1148}{0.1172} & \rs{0.8539}{0.0209} \\
\quad Gaussian Copula
& \rs{0.9655}{0.0000} & \rs{0.9618}{0.0000} & \rs{0.9338}{0.0000} & \rs{0.9793}{0.0000} & \rsu{0.8571}{0.0000} & \rsu{0.7333}{0.0000} & \rs{0.9427}{0.0000}
& \rs{0.8874}{0.0000} & \rs{0.8695}{0.0000} & \rs{0.7801}{0.0000} & \rs{0.9159}{0.0000} & \rs{0.4150}{0.0000} & \rsu{0.7895}{0.0000} & \rs{0.7081}{0.0000} \\
\quad K-IPO
& \rsb{0.9812}{0.0016} & \rsb{0.9777}{0.0019} & \rsb{0.9608}{0.0034} & \rsb{0.9873}{0.0012} & \rsb{0.8666}{0.0301} & \rsb{1.0000}{0.0000} & \rsb{0.9828}{0.0007}
& \rsu{0.9653}{0.0021} & \rsu{0.9603}{0.0027} & \rs{0.9308}{0.0048} & \rsu{0.9778}{0.0018} & \rsb{0.4640}{0.0156} & \rsb{0.9095}{0.0263} & \rsb{0.9393}{0.0032} \\
\quad SMOTE
& \rsu{0.9734}{0.0027} & \rsu{0.9688}{0.0030} & \rsu{0.9451}{0.0053} & \rsu{0.9820}{0.0015} & \rs{0.7301}{0.0280} & \rs{0.6000}{0.0000} & \rsu{0.9770}{0.0008}
& \rs{0.9528}{0.0017} & \rs{0.9456}{0.0021} & \rs{0.9050}{0.0036} & \rs{0.9690}{0.0012} & \rsu{0.4449}{0.0121} & \rs{0.7810}{0.0317} & \rsu{0.9264}{0.0013} \\
\quad TabDDPM
& \rs{0.8730}{0.0420} & \rs{0.8558}{0.0482} & \rs{0.7492}{0.0888} & \rs{0.9060}{0.0390} & \rs{0.7810}{0.0751} & \rs{0.4667}{0.2266} & \rs{0.8330}{0.0549}
&  \rs{0.9639}{0.0055} & \rs{0.9599}{0.0061} & \rsu{0.9310}{0.0103} & \rs{0.9766}{0.0039} & \rs{0.4061}{0.0291} & \rs{0.0726}{0.1578} & \rs{0.9200}{0.0115} \\
\quad TVAE
& \rs{0.9458}{0.0044} & \rs{0.9365}{0.0050} & \rs{0.8877}{0.0090} & \rs{0.9628}{0.0032} & \rs{0.6730}{0.0778} & \rs{0.4400}{0.0562} & \rs{0.9125}{0.0042}
&\rs{0.9645}{0.0016} & \rs{0.9601}{0.0021} & \rs{0.9307}{0.0040} & \rs{0.9769}{0.0010} & \rs{0.4082}{0.0187} & \rs{0.4684}{0.0181} & \rs{0.9008}{0.0043} \\
\cmidrule(lr){1-15}

& \multicolumn{7}{l||}{\textit{D2}} & \multicolumn{7}{l}{\textit{D7}} \\
\quad CTGAN
& \rsb{0.9320}{0.0027} & \rsb{0.9251}{0.0037} & \rsb{0.8759}{0.0084} & \rsb{0.9587}{0.0022} & \rs{0.8175}{0.0259} & \rs{0.2572}{0.2054} & \rs{0.8969}{0.0051}
& \rs{0.9095}{0.0064} & \rs{0.8975}{0.0079} & \rs{0.8248}{0.0153} & \rsu{0.9404}{0.0058} & \rsu{0.4809}{0.0322} & \rs{0.3083}{0.1028} & \rs{0.8083}{0.0370} \\
\quad Gaussian Copula
& \rs{0.8436}{0.0000} & \rs{0.8238}{0.0000} & \rs{0.6803}{0.0000} & \rs{0.8497}{0.0000} & \rs{0.8095}{0.0000} & \rsu{0.8571}{0.0000} & \rs{0.7301}{0.0000}
& \rs{0.9073}{0.0000} & \rs{0.8957}{0.0000} & \rs{0.8219}{0.0000} & \rs{0.9382}{0.0000} & \rs{0.4667}{0.0000} & \rsu{0.7667}{0.0000} & \rs{0.7796}{0.0000} \\
\quad K-IPO
& \rs{0.9008}{0.0058} & \rs{0.8863}{0.0062} & \rs{0.7961}{0.0115} & \rs{0.9002}{0.0042} & \rsb{0.8810}{0.0242} & \rsb{1.0000}{0.0000} & \rsb{0.9192}{0.0018}
& \rsu{0.9166}{0.0048} & \rsu{0.9051}{0.0056} & \rsu{0.8349}{0.0105} & \rs{0.9390}{0.0039} & \rs{0.4753}{0.0239} & \rsb{0.8400}{0.0238} & \rsb{0.9069}{0.0022} \\
\quad SMOTE
& \rs{0.8940}{0.0048} & \rs{0.8788}{0.0051} & \rs{0.7821}{0.0092} & \rs{0.8940}{0.0035} & \rsu{0.8444}{0.0146} & \rsu{0.8571}{0.0000} & \rsu{0.9045}{0.0024}
& \rs{0.9041}{0.0026} & \rs{0.8896}{0.0029} & \rs{0.8040}{0.0057} & \rs{0.9237}{0.0031} & \rs{0.4457}{0.0219} & \rs{0.6933}{0.0353} & \rs{0.8937}{0.0018} \\
\quad TabDDPM
& \rs{0.8669}{0.0351} & \rs{0.8482}{0.0423} & \rs{0.7487}{0.0776} & \rs{0.9051}{0.0309} & \rs{0.8270}{0.0272} & \rs{-0.0214}{0.1013} & \rs{0.8464}{0.0328}
& \rsb{0.9176}{0.0016} & \rsb{0.9073}{0.0022} & \rsb{0.8428}{0.0063} & \rsb{0.9473}{0.0015} & \rsb{0.4905}{0.0299} & \rs{0.2717}{0.0933} & \rsu{0.9036}{0.0026}  \\
\quad TVAE
& \rsu{0.9186}{0.0028} & \rsu{0.9074}{0.0030} & \rsu{0.8394}{0.0051} & \rsu{0.9470}{0.0019} & \rs{0.7595}{0.0195} & \rs{0.4714}{0.0904} & \rs{0.8647}{0.0047}
& \rs{0.9145}{0.0016} & \rs{0.9027}{0.0020} & \rs{0.8311}{0.0045} & \rs{0.9395}{0.0020} & \rs{0.3867}{0.0247} & \rs{0.2700}{0.0422} & \rs{0.8746}{0.0042} \\
\cmidrule(lr){1-15}

& \multicolumn{7}{l||}{\textit{D3}} & \multicolumn{7}{l}{\textit{D8}} \\
\quad CTGAN
& \rs{0.7904}{0.0132} & \rs{0.7616}{0.0136} & \rs{0.5799}{0.0280} & \rs{0.8189}{0.0173} & \rsb{0.8114}{0.0481} & \rs{0.4857}{0.1118} & \rs{0.6932}{0.0208}
& \rs{0.9273}{0.0080} & \rs{0.9193}{0.0098} & \rs{0.8650}{0.0178} & \rs{0.9487}{0.0058} & \rs{0.7043}{0.0639} & \rs{0.4515}{0.1542} & \rs{0.8006}{0.0208} \\
\quad Gaussian Copula
& \rs{0.7688}{0.0000} & \rs{0.7427}{0.0000} & \rs{0.5352}{0.0000} & \rs{0.7923}{0.0000} & \rs{0.7571}{0.0000} & \rsu{0.7143}{0.0000} & \rs{0.6627}{0.0000}
& \rs{0.9065}{0.0000} & \rs{0.8941}{0.0000} & \rs{0.8162}{0.0000} & \rs{0.9335}{0.0000} & \rsu{0.7714}{0.0000} & \rsu{0.5758}{0.0000} & \rs{0.7427}{0.0000} \\
\quad K-IPO
& \rs{0.8113}{0.0093} & \rs{0.7877}{0.0113} & \rs{0.6219}{0.0188} & \rs{0.8271}{0.0111} & \rsu{0.7705}{0.0240} & \rsb{0.9524}{0.0502} & \rsu{0.8253}{0.0138}
& \rsb{0.9467}{0.0041} & \rsb{0.9400}{0.0047} & \rsb{0.8965}{0.0083} & \rsb{0.9648}{0.0028} & \rs{0.6714}{0.0270} & \rsb{0.8364}{0.0256} & \rsb{0.9082}{0.0028} \\
\quad SMOTE
& \rs{0.7964}{0.0039} & \rs{0.7706}{0.0032} & \rs{0.5900}{0.0090} & \rs{0.8106}{0.0040} & \rs{0.7467}{0.0281} & \rs{0.7143}{0.0449} & \rs{0.8025}{0.0031}
& \rsu{0.9346}{0.0061} & \rsu{0.9256}{0.0075} & \rsu{0.8708}{0.0141} & \rsu{0.9548}{0.0047} & \rs{0.6285}{0.0530} & \rs{0.3303}{0.0390} & \rsu{0.8840}{0.0026} \\
\quad TabDDPM
& \rsb{0.9315}{0.0002} & \rsb{0.9264}{0.0002} & \rsb{0.8845}{0.0004} & \rsb{0.9488}{0.0004} & \rs{0.7352}{0.0378} & \rs{-0.2857}{0.1634} & \rsb{0.9022}{0.0008}
& \rs{0.9126}{0.0267} & \rs{0.9026}{0.0320} & \rs{0.8425}{0.0514} & \rs{0.9374}{0.0203} & \rsb{0.7743}{0.0598} & \rs{0.3606}{0.1233} & \rs{0.8613}{0.0452} \\
\quad TVAE
& \rsu{0.8381}{0.0042} & \rsu{0.8158}{0.0052} & \rsu{0.6783}{0.0086} & \rsu{0.8670}{0.0038} & \rs{0.6705}{0.0295} & \rs{0.4381}{0.1821} & \rs{0.7934}{0.0041}
& \rs{0.9076}{0.0104} & \rs{0.8954}{0.0123} & \rs{0.8223}{0.0225} & \rs{0.9333}{0.0071} & \rs{0.4357}{0.0422} & \rs{-0.0697}{0.0249} & \rs{0.8170}{0.0126} \\
\cmidrule(lr){1-15}

& \multicolumn{7}{l||}{\textit{D4}} & \multicolumn{7}{l}{\textit{D9}} \\
\quad CTGAN
& \rsu{0.9521}{0.0016} & \rsu{0.9485}{0.0019} & \rsu{0.9141}{0.0034} & \rsu{0.9672}{0.0012} & \rs{0.7205}{0.0287} & \rs{0.3752}{0.1253} & \rs{0.8677}{0.0111}
& \rsu{0.9432}{0.0031} & \rsu{0.9363}{0.0037} & \rsu{0.8902}{0.0069} & \rsu{0.9649}{0.0025} & \rs{0.3548}{0.0545} & \rs{0.4177}{0.2129} & \rs{0.8550}{0.0197} \\
\quad Gaussian Copula
& \rs{0.9383}{0.0000} & \rs{0.9320}{0.0000} & \rs{0.8859}{0.0000} & \rs{0.9568}{0.0000} & \rs{0.7190}{0.0000} & \rs{0.6190}{0.0000} & \rs{0.8405}{0.0000}
& \rs{0.9332}{0.0000} & \rs{0.9250}{0.0000} & \rs{0.8709}{0.0000} & \rs{0.9571}{0.0000} & \rs{0.4683}{0.0000} & \rs{0.4853}{0.0000} & \rs{0.8151}{0.0000} \\
\quad K-IPO
& \rs{0.9046}{0.0061} & \rs{0.8909}{0.0073} & \rs{0.8073}{0.0135} & \rs{0.9227}{0.0066} & \rsb{0.7667}{0.0087} & \rsb{0.8400}{0.0339} & \rsu{0.9069}{0.0057}
& \rs{0.9411}{0.0022} & \rs{0.9315}{0.0023} & \rs{0.8791}{0.0038} & \rs{0.9558}{0.0018} & \rsb{0.5682}{0.0412} & \rsb{0.8493}{0.0388} & \rsb{0.9319}{0.0037} \\
\quad SMOTE
& \rs{0.8904}{0.0040} & \rs{0.8745}{0.0048} & \rs{0.7759}{0.0088} & \rs{0.9035}{0.0037} & \rsu{0.7510}{0.0194} & \rsu{0.7257}{0.0326} & \rs{0.8922}{0.0049}
& \rs{0.9360}{0.0028} & \rs{0.9254}{0.0030} & \rs{0.8681}{0.0055} & \rs{0.9516}{0.0020} & \rsu{0.5349}{0.0443} & \rsu{0.6822}{0.0408} & \rsu{0.9245}{0.0025} \\
\quad TabDDPM
& \rsb{0.9540}{0.0025} & \rsb{0.9517}{0.0028} & \rsb{0.9213}{0.0041} & \rsb{0.9688}{0.0017} & \rs{0.7181}{0.0359} & \rs{0.0248}{0.0739} & \rsb{0.9294}{0.0016}
& \rsb{0.9495}{0.0007} & \rsb{0.9436}{0.0010} & \rsb{0.9030}{0.0025} & \rsb{0.9689}{0.0006} & \rs{0.5016}{0.1058} & \rs{0.2956}{0.0910} & \rs{0.9158}{0.0017} \\
\quad TVAE
& \rs{0.9394}{0.0040} & \rs{0.9334}{0.0044} & \rs{0.8880}{0.0068} & \rs{0.9592}{0.0026} & \rs{0.6843}{0.0260} & \rs{0.2152}{0.0465} & \rs{0.8708}{0.0039}
& \rs{0.9395}{0.0006} & \rs{0.9319}{0.0008} & \rs{0.8824}{0.0018} & \rs{0.9603}{0.0008} & \rs{0.4143}{0.0282} & \rs{-0.0073}{0.0915} & \rs{0.9088}{ 0.0038} \\
\cmidrule(lr){1-15}

& \multicolumn{7}{l||}{\textit{D5}} & \multicolumn{7}{l}{\textit{D10}} \\
\quad CTGAN
& \rs{0.9130}{0.0104} & \rs{0.9022}{0.0124} & \rs{0.8350}{0.0213} & \rs{0.9391}{0.0091} & \rs{0.4417}{0.0440} & \rs{0.1200}{0.3479} & \rs{0.7877}{0.0278}
& \rs{0.9851}{0.0005} & \rs{0.9805}{0.0007} & \rs{0.9659}{0.0012} & \rs{0.9833}{0.0020} & \rs{0.8548}{0.0404} & \rsu{0.9600}{0.0644} & \rs{0.9678}{0.0026} \\
\quad Gaussian Copula
& \rsu{0.9286}{0.0000} & \rsu{0.9203}{0.0000} & \rsu{0.8643}{0.0000} & \rsu{0.9525}{0.0000} & \rsu{0.5476}{0.0000} & \rs{0.4667}{0.0000} & \rs{0.7904}{0.0000}
& \rs{0.9850}{0.0000} & \rs{0.9804}{0.0000} & \rs{0.9657}{0.0000} & \rs{0.9842}{0.0000} & \rs{0.8214}{0.0000} & \rsb{1.0000}{0.0000} & \rs{0.9678}{0.0000} \\
\quad K-IPO
& \rs{0.9165}{0.0072} & \rs{0.9044}{0.0084} & \rs{0.8312}{0.0157} & \rs{0.9377}{0.0064} & \rsb{0.5559}{0.0195} & \rsb{0.9511}{0.0532} & \rsu{0.9139}{0.0060}
& \rsb{1.0000}{0.0000} & \rsb{1.0000}{0.0000} & \rsb{1.0000}{0.0000} & \rsb{1.0000}{0.0000} & \rsb{0.8595}{0.0431} & \rsb{1.0000}{0.0000} & \rsb{0.9881}{0.0013} \\
\quad SMOTE
& \rs{0.8991}{0.0031} & \rs{0.8843}{0.0037} & \rs{0.7947}{0.0076} & \rs{0.9233}{0.0026} & \rs{0.5357}{0.0317} & \rsu{0.7689}{0.0459} & \rs{0.9039}{0.0027}
& \rsb{1.0000}{0.0000} & \rsb{1.0000}{0.0000} & \rsb{1.0000}{0.0000} & \rsb{1.0000}{0.0000} & \rs{0.8214}{0.0000} & \rs{0.7333}{0.0000} & \rsu{0.9871}{0.0000} \\
\quad TabDDPM
& \rsb{0.9493}{0.0015} & \rsb{0.9453}{0.0017} & \rsb{0.9088}{0.0027} & \rsb{0.9669}{0.0008} & \rs{0.5333}{0.0411} & \rs{-0.1467}{0.2341} & \rsb{0.9216}{0.0017}
& \rs{0.4770}{0.0164} & \rs{0.4328}{0.0179} & \rs{-0.0455}{0.0327} & \rs{0.4579}{0.0158} & \rs{0.6000}{0.0373} & \rs{0.4533}{0.2626} & \rs{0.5983}{0.0122} \\
\quad TVAE
& \rs{0.9257}{0.0030} & \rsu{0.9163}{0.0033} & \rs{0.8543}{0.0059} & \rs{0.9420}{0.0027} & \rs{0.4798}{0.0416} & \rs{0.4533}{0.1222} & \rs{0.8404}{0.0058}
& \rsu{0.9996}{0.0006} & \rsu{0.9995}{0.0006} & \rsu{0.9991}{0.0011} & \rsu{0.9998}{0.0003} & \rsu{0.8464}{0.0379} & \rs{0.8800}{0.0422} & \rs{0.9846}{0.0007} \\
\bottomrule
\end{tabular}%
}
\end{sidewaystable}

\begin{sidewaystable}[p]
\caption{%
Comparison of synthetic data generation methods on datasets D11--D20 across seven evaluation metrics.
Entries represent mean \texorpdfstring{$\pm$}{±} standard deviation over multiple runs.
\textbf{Bold} values indicate the best performance per metric per dataset.
}
\label{tab:results_d11_d20}
\centering
\scriptsize
\setlength{\tabcolsep}{3pt}
\renewcommand{\arraystretch}{1.1}

\resizebox{\linewidth}{!}{%
\begin{tabular}{l||ccccccc||ccccccc}
\toprule
& \multicolumn{7}{c||}{\textbf{Datasets D11--D15}}
& \multicolumn{7}{c}{\textbf{Datasets D16--D20}} \\
\cmidrule(lr){2-8}\cmidrule(lr){9-15}
\textbf{Method}
& \textbf{Bal.\ Acc.} & \textbf{F1} & \textbf{MCC} & $\boldsymbol{P}$ & $\boldsymbol{I}$ & \textbf{Kendall's} $\boldsymbol{\tau}$ & $\boldsymbol{1\!-\!N3}$
& \textbf{Bal.\ Acc.} & \textbf{F1} & \textbf{MCC} & $\boldsymbol{P}$ & $\boldsymbol{I}$ & \textbf{Kendall's} $\boldsymbol{\tau}$ & $\boldsymbol{1\!-\!N3}$ \\
\midrule

& \multicolumn{7}{l||}{\textit{D11}} & \multicolumn{7}{l}{\textit{D16}} \\
\quad CTGAN
& \rs{0.9819}{0.0049} & \rs{0.9791}{0.0054} & \rs{0.9634}{0.0094} & \rs{0.9896}{0.0029} & \rs{0.5651}{0.0609} & \rs{0.5867}{0.2127} & \rs{0.9696}{0.0096}
& \rs{0.8976}{0.0082} & \rs{0.8840}{0.0093} & \rs{0.8022}{0.0146} & \rs{0.9277}{0.0073} & \rs{0.3270}{0.0814} & \rs{0.1827}{0.0825} & \rs{0.7472}{0.0335} \\
\quad Gaussian Copula
& \rs{0.9806}{0.0000} & \rs{0.9782}{0.0000} & \rs{0.9619}{0.0000} & \rs{0.9891}{0.0000} & \rs{0.4444}{0.0000} & \rs{0.6000}{0.0000} & \rs{0.9617}{0.0000}
& \rs{0.8916}{0.0000} & \rs{0.8771}{0.0000} & \rs{0.7841}{0.0000} & \rs{0.9194}{0.0000} & \rs{0.2857}{0.0000} & \rs{0.4113}{0.0000} & \rs{0.7389}{0.0000} \\
\quad K-IPO
& \rsb{0.9873}{0.0009} & \rsb{0.9842}{0.0010} & \rsb{0.9723}{0.0018} & \rsb{0.9905}{0.0005} & \rsb{0.6206}{0.0117} & \rsb{1.0000}{0.0000} & \rsb{0.9947}{0.0001}
& \rsb{0.9879}{0.0015} & \rsb{0.9854}{0.0019} & \rsb{0.9743}{0.0033} & \rsb{0.9843}{0.0013} & \rsb{0.4746}{0.0439} & \rsb{0.7746}{0.0082} & \rsb{0.9543}{0.0005} \\
\quad SMOTE
& \rsu{0.9868}{0.0007} & \rsu{0.9835}{0.0008} & \rsu{0.9710}{0.0014} & \rs{0.9902}{0.0004} & \rs{0.5905}{0.0222} & \rsu{0.8667}{0.0000} & \rsu{0.9946}{0.0000}
& \rsu{0.9609}{0.0099} & \rsu{0.9549}{0.0112} & \rsu{0.9209}{0.0196} & \rsu{0.9719}{0.0047} & \rsu{0.4508}{0.0353} & \rsu{0.5544}{0.0981} & \rs{0.8499}{0.0382} \\
\quad TabDDPM
& \rs{0.9481}{0.0117} & \rs{0.9416}{0.0128} & \rs{0.8993}{0.0206} & \rs{0.9639}{0.0091} & \rsu{0.6127}{0.0250} & \rs{0.4933}{0.1968} & \rs{0.9206}{0.0520}
& \rs{0.8347}{0.0196} & \rs{0.8110}{0.0222} & \rs{0.6715}{0.0413} & \rs{0.8708}{0.0217} & \rs{0.3572}{0.0418} & \rs{0.2105}{0.1528} & \rs{0.6878}{0.0311} \\
\quad TVAE
& \rs{0.9857}{0.0007} & \rs{0.9827}{0.0009} & \rs{0.9697}{0.0016} & \rsu{0.9904}{0.0005} & \rs{0.5397}{0.0225} & \rs{0.7466}{0.0422} & \rs{0.9728}{0.0007}
& \rs{0.9253}{0.0034} & \rs{0.9157}{0.0037} & \rs{0.8553}{0.0062} & \rs{0.9511}{0.0021} & \rs{0.2127}{0.0383} & \rs{0.1532}{0.0562} & \rsu{0.8670}{0.0084} \\
\cmidrule(lr){1-15}

& \multicolumn{7}{l||}{\textit{D12}} & \multicolumn{7}{l}{\textit{D17}} \\
\quad CTGAN
& \rsb{0.9572}{0.0013} & \rsb{0.9527}{0.0017} & \rsb{0.9188}{0.0033} & \rsb{0.9742}{0.0009} & \rs{0.5817}{0.0174} & \rs{0.0211}{0.0802} & \rs{0.9374}{0.0032}
& \rsu{0.9908}{0.0001} & \rsu{0.9887}{0.0000} & \rsu{0.9802}{0.0000} & \rsu{0.9938}{0.0002} & \rsb{0.6508}{0.0000} & \rsb{1.0000}{0.0000} & \rsb{0.9953}{0.0000} \\
\quad Gaussian Copula
& \rsu{0.9568}{0.0000} & \rsu{0.9519}{0.0000} & \rsu{0.9168}{0.0000} & \rsu{0.9736}{0.0000} & \rs{0.6270}{0.0000} & \rs{0.5474}{0.0000} & \rs{0.9031}{0.0000}
& \rs{0.9648}{0.0000} & \rs{0.9592}{0.0000} & \rs{0.9287}{0.0000} & \rs{0.9778}{0.0000} & \rs{0.6349}{0.0000} & \rsb{1.0000}{0.0000} & \rs{0.9187}{0.0000} \\
\quad K-IPO
& \rs{0.9488}{0.0067} & \rs{0.9412}{0.0072} & \rs{0.8962}{0.0125} & \rs{0.9620}{0.0027} & \rs{0.6190}{0.0109} & \rsb{0.8105}{0.0489} & \rsb{0.9547}{0.0111}
& \rsb{0.9918}{0.0011} & \rsb{0.9898}{0.0012} & \rsb{0.9821}{0.0021} & \rsb{0.9945}{0.0007} & \rsb{0.6508}{0.0000} & \rsb{1.0000}{0.0000} & \rsu{0.9950}{0.0003} \\
\quad SMOTE
& \rs{0.9376}{0.0024} & \rs{0.9284}{0.0028} & \rs{0.8737}{0.0049} & \rs{0.9552}{0.0013} & \rs{0.5972}{0.0073} & \rsu{0.7274}{0.0376} & \rsu{0.9402}{0.0021}
& \rs{0.9905}{0.0003} & \rs{0.9884}{0.0004} & \rs{0.9796}{0.0006} & \rsu{0.9938}{0.0003} & \rsb{0.6508}{0.0000} & \rsb{1.0000}{0.0000} & \rsb{0.9953}{0.0000} \\
\quad TabDDPM
& \rs{0.9181}{0.0041} & \rs{0.9067}{0.0047} & \rs{0.8387}{0.0075} & \rs{0.9443}{0.0031} & \rsb{0.6377}{0.0123} & \rs{-0.0716}{0.1492} & \rs{0.8686}{0.0103}
& \rs{0.9767}{0.0019} & \rs{0.9720}{0.0021} & \rs{0.9509}{0.0037} & \rs{0.9851}{0.0013} & \rsu{0.6317}{0.0123} & \rs{0.5000}{0.2539} & \rs{0.8827}{0.0325} \\
\quad TVAE
& \rs{0.9455}{0.0014} & \rs{0.9385}{0.0019} & \rs{0.8930}{0.0037} & \rs{0.9656}{0.0010} & \rsu{0.6329}{0.0111} & \rs{0.3968}{0.0884} & \rs{0.9106}{0.0026}
& \rs{0.9376}{0.0085} & \rs{0.9292}{0.0100} & \rs{0.8769}{0.0178} & \rs{0.9600}{0.0058} & \rs{0.6190}{0.0146} & \rsu{0.9800}{0.0632} & \rs{0.8810}{0.0061} \\
\cmidrule(lr){1-15}

& \multicolumn{7}{l||}{\textit{D13}} & \multicolumn{7}{l}{\textit{D18}} \\
\quad CTGAN
& \rs{0.8830}{0.0109} & \rs{0.8687}{0.0120} & \rs{0.7779}{0.0239} & \rs{0.9074}{0.0111} & \rs{0.5547}{0.0413} & \rs{0.2857}{0.1918} & \rs{0.7915}{0.0268}
& \rs{0.9953}{0.0003} & \rs{0.9942}{0.0004} & \rs{0.9898}{0.0006} & \rs{0.9974}{0.0002} & \rsu{0.6415}{0.0078} & \rs{0.7233}{0.0081} & \rs{0.9314}{0.0137} \\
\quad Gaussian Copula
& \rsb{0.9304}{0.0000} & \rsb{0.9232}{0.0000} & \rsb{0.8726}{0.0000} & \rsb{0.9515}{0.0000} & \rs{0.5357}{0.0000} & \rsu{0.9048}{0.0000} & \rsu{0.8647}{0.0000}
& \rs{0.9951}{0.0000} & \rs{0.9940}{0.0000} & \rs{0.9893}{0.0000} & \rs{0.9971}{0.0000} & \rs{0.6355}{0.0000} & \rs{0.7785}{0.0000} & \rs{0.9756}{0.0000} \\
\quad K-IPO
& \rsu{0.9030}{0.0060} & \rsu{0.8909}{0.0065} & \rsu{0.8052}{0.0142} & \rsu{0.9123}{0.0077} & \rsb{0.7500}{0.0202} & \rsb{0.9714}{0.0460} & \rsb{0.9279}{0.0042}
& \rsb{0.9981}{0.0003} & \rsb{0.9974}{0.0004} & \rsb{0.9955}{0.0007} & \rsb{0.9985}{0.0002} & \rsb{0.6434}{0.0036} & \rsb{0.8509}{0.0059} & \rsu{0.9885}{0.0004} \\
\quad SMOTE
& \rs{0.8776}{0.0095} & \rs{0.8615}{0.0108} & \rs{0.7582}{0.0181} & \rs{0.8918}{0.0079} & \rsu{0.7226}{0.0313} & \rs{0.7238}{0.0301} & \rsb{0.9279}{0.0045}
& \rsu{0.9979}{0.0003} & \rsu{0.9972}{0.0004} & \rsu{0.9951}{0.0008} & \rsu{0.9984}{0.0002} & \rs{0.6369}{0.0051} & \rsu{0.8309}{0.0027} & \rsb{0.9890}{0.0001} \\
\quad TabDDPM
& \rs{0.7353}{0.0167} & \rs{0.6997}{0.0228} & \rs{0.4703}{0.0314} & \rs{0.7612}{0.0177} & \rs{0.6976}{0.0347} & \rs{0.5428}{0.1405} & \rs{0.6413}{0.0124}
& \rs{0.9951}{0.0009} & \rs{0.9941}{0.0010} & \rs{0.9897}{0.0017} & \rs{0.9973}{0.0005} & \rs{0.6310}{0.0090} & \rs{0.1289}{0.0575} & \rs{0.7126}{0.0569} \\
\quad TVAE
& \rs{0.8554}{0.0077} & \rs{0.8393}{0.0085} & \rs{0.7188}{0.0148} & \rs{0.8667}{0.0086} & \rs{0.4774}{0.0467} & \rs{0.6190}{0.0000} & \rs{0.6988}{0.0124}
& \rs{0.9968}{0.0006} & \rs{0.9961}{0.0008} & \rs{0.9932}{0.0014} & \rs{0.9982}{0.0004} & \rs{0.6371}{0.0061} & \rs{0.7728}{0.0080} & \rs{0.9869}{0.0001} \\
\cmidrule(lr){1-15}

& \multicolumn{7}{l||}{\textit{D14}} & \multicolumn{7}{l}{\textit{D19}} \\
\quad CTGAN
& \rsb{0.9652}{0.0011} & \rsb{0.9618}{0.0011} & \rsb{0.9343}{0.0019} & \rsb{0.9789}{0.0006} & \rs{0.4976}{0.0689} & \rs{0.2534}{0.1929} & \rs{0.9393}{0.0034}
& \rs{0.9895}{0.0030} & \rs{0.9878}{0.0037} & \rs{0.9786}{0.0065} & \rs{0.9943}{0.0018} & \rsu{0.6016}{0.0513} & \rs{0.4484}{0.1170} & \rs{0.9865}{0.0056} \\
\quad Gaussian Copula
& \rsu{0.9507}{0.0000} & \rsu{0.9442}{0.0000} & \rsu{0.9029}{0.0000} & \rsu{0.9687}{0.0000} & \rsb{0.5714}{0.0000} & \rsb{1.0000}{0.0000} & \rs{0.9159}{0.0000}
& \rs{0.9858}{0.0000} & \rs{0.9838}{0.0000} & \rs{0.9716}{0.0000} & \rs{0.9923}{0.0000} & \rs{0.5556}{ 0.0000} & \rs{0.9121}{0.0000} & \rs{0.9910}{0.0000} \\
\quad K-IPO
& \rs{0.9378}{0.0074} & \rs{0.9291}{0.0087} & \rs{0.8765}{0.0155} & \rs{0.9591}{0.0052} & \rs{0.5119}{0.0194} & \rsb{1.0000}{0.0000} & \rsb{0.9509}{0.0067}
& \rsu{0.9963}{0.0005} & \rsu{0.9958}{0.0006} & \rsu{0.9926}{0.0010} & \rsu{0.9981}{0.0003} & \rsb{0.6071}{0.0180} & \rsb{0.9626}{0.0208} & \rsu{0.9980}{0.0002} \\
\quad SMOTE
& \rs{0.9363}{0.0088} & \rs{0.9274}{0.0102} & \rs{0.8735}{0.0179} & \rs{0.9582}{0.0063} & \rs{0.4940}{0.0369} & \rsu{0.9689}{0.0366} & \rsu{0.9506}{0.0068}
& \rsb{0.9975}{0.0006} & \rsb{0.9970}{0.0006} & \rsb{0.9947}{0.0011} & \rsb{0.9986}{0.0003} & \rs{0.5968}{0.0336} & \rsu{0.9341}{0.0000} & \rsb{0.9984}{0.0001} \\
\quad TabDDPM
& \rs{0.8169}{0.1286} & \rs{0.7905}{0.1483} & \rs{0.6374}{0.2593} & \rs{0.8451}{0.1333} & \rs{0.5536}{0.0470} & \rs{0.7511}{0.0843} & \rs{0.7643}{0.1213}
& \rs{0.9879}{0.0018} & \rs{0.9860}{0.0019} & \rs{0.9756}{0.0033} & \rs{0.9935}{0.0009} & \rs{0.6008}{0.0495} & \rs{0.6989}{0.1294} & \rs{0.9891}{0.0013} \\
\quad TVAE
& \rs{0.9383}{0.0017} & \rs{0.9299}{0.0019} & \rs{0.8779}{0.0033} & \rs{0.9592}{0.0013} & \rsu{0.5547}{0.0556} & \rs{0.9156}{0.0328} & \rs{0.8603}{0.0041}
& \rs{0.9885}{0.0014} & \rs{0.9865}{0.0018} & \rs{0.9763}{0.0032} & \rs{0.9937}{0.0009} & \rs{0.5913}{0.0273} & \rs{0.8549}{0.0362} & \rs{0.9898}{0.0007} \\
\cmidrule(lr){1-15}

& \multicolumn{7}{l||}{\textit{D15}} & \multicolumn{7}{l}{\textit{D20}} \\
\quad CTGAN
& \rs{0.9287}{0.0134} & \rs{0.9201}{0.0153} & \rs{0.8643}{0.0248} & \rs{0.9532}{0.0102} & \rs{0.5143}{0.1010} & \rs{0.3378}{0.1820} & \rs{0.8203}{0.0363}
& \rs{0.9977}{0.0004} & \rs{0.9975}{0.0004} & \rs{0.9956}{0.0007} & \rs{0.9988}{0.0002} & \rs{0.6323}{0.0311} & \rs{0.4083}{0.1550} & \rs{0.9928}{0.0010} \\
\quad Gaussian Copula
& \rs{0.8681}{0.0000} & \rs{0.8488}{0.0000} & \rs{0.7555}{0.0000} & \rs{0.9068}{0.0000} & \rs{0.6508}{0.0000} & \rs{0.7778}{0.0000} & \rs{0.7734}{0.0000}
& \rs{0.9981}{0.0000} & \rs{0.9977}{0.0000} & \rs{0.9960}{0.0000} & \rs{0.9990}{0.0000} & \rs{0.6349}{0.0000} & \rs{0.6500}{0.0000} & \rs{0.9979}{0.0000} \\
\quad K-IPO
& \rsb{0.9751}{0.0038} & \rsb{0.9719}{0.0047} & \rsb{0.9509}{0.0084} & \rsb{0.9858}{0.0024} & \rsb{0.6746}{0.0135} & \rsb{0.9556}{0.0363} & \rsb{0.9599}{0.0013}
& \rsb{1.0000}{0.0000} & \rsb{1.0000}{0.0000} & \rsb{0.9999}{0.0001} & \rsb{1.0000}{0.0000} & \rsb{0.6534}{0.0128} & \rsb{0.9350}{0.0095} & \rsu{0.9992}{0.0001} \\
\quad SMOTE
& \rs{0.9703}{0.0013} & \rs{0.9658}{0.0013} & \rs{0.9401}{0.0023} & \rs{0.9827}{0.0007} & \rs{0.6619}{0.0107} & \rsu{0.9067}{0.0389} & \rsu{0.9589}{0.0013}
& \rsu{0.9999}{0.0001} & \rsu{0.9999}{0.0001} & \rsb{0.9999}{0.0001} & \rsb{1.0000}{0.0000} & \rs{0.6381}{0.0087} & \rs{0.9050}{0.0369} & \rsb{0.9993}{0.0001} \\
\quad TabDDPM
& \rs{0.8839}{0.0045} & \rs{0.8676}{0.0052} & \rs{0.7739}{0.0091} & \rs{0.9184}{0.0034} & \rs{0.6603}{0.0111} & \rs{0.8133}{0.1064} & \rs{0.7925}{0.0033}
& \rs{0.9909}{0.0010} & \rs{0.9899}{0.0012} & \rs{0.9823}{0.0021} & \rs{0.9952}{0.0006} & \rsu{0.6503}{0.0326} & \rs{0.6817}{0.1026} & \rs{0.9806}{0.0021} \\
\quad TVAE
& \rsu{0.9723}{0.0014} & \rsu{0.9696}{0.0016} & \rsu{0.9476}{0.0028} & \rsu{0.9844}{0.0009} & \rsu{0.6699}{0.0164} & \rs{0.5556}{0.0983} & \rs{0.9441}{0.0016}
& \rs{0.9979}{0.0006} & \rs{0.9976}{0.0006} & \rs{0.9959}{0.0011} & \rsu{0.9989}{0.0003} & \rs{0.6455}{0.0112} & \rsu{0.9083}{0.0226} & \rs{0.9963}{0.0005} \\
\bottomrule
\end{tabular}%
}
\end{sidewaystable}

The results indicate that K-IPO offers the most favorable trade-off between predictive performance and interpretability. Despite the structural constraints imposed during augmentation, K-IPO remains highly competitive, achieving the highest number of wins in Balanced Accuracy and leading in F1-score, MCC, and $P$ (9 wins each). These findings suggest that the sample-selection mechanism preserves the downstream classifier utility by filtering out synthetic instances that could distort the decision boundary. However, this advantage was not uniform across all datasets, as expected, given the restrictive acceptance criterion. In some cases, unconstrained generators, such as CTGAN or TabDDPM, achieved higher predictive scores; however, these gains often came at the cost of weaker feature importance preservation. For instance, in D12, CTGAN obtained the best predictive performance, but its Kendall’s $\tau$ was only 0.0211, indicating a severe distortion of the original feature importance ranking. Gaussian Copula showed a stable but weaker predictive performance, achieving only one win in Balanced Accuracy, F1-score, MCC, and $P$. This suggests that its lightweight statistical fit-and-sample mechanism can be effective in isolated cases, but its distributional assumptions may limit its ability to capture dataset-specific minority class structures. Moreover, the zero standard deviations reported for the Gaussian Copula arise from the design of the Synthetic Data Vault (SDV) library~\cite{SDV}, which was used to implement this generator. Specifically, the SDV API does not expose the random state used during sample generation and instead relies on its internal default random state. Because the same train/test split was used across all experimental runs, the generator produced identical augmented datasets in every repetition. Consequently, the evaluation metrics remained unchanged throughout the runs, leading to a zero standard deviation.

The interpretability results provide stronger evidence for the main contribution of K-IPO. K-IPO achieves the best or tied-best Kendall’s $\tau$ on all 20 datasets, substantially outperforming other methods in terms of feature importance rank preservation. It also achieves the highest explainability consistency score $I$ in 15 out of 20 datasets, whereas the closest competitor, TabDDPM, achieves only 3 wins. These results support the main premise of this study: synthetic samples that appear useful from a predictive or distributional perspective may still alter the feature importance structure underlying model decisions when no explicit importance preservation constraint is imposed. In contrast, K-IPO directly controls the source of the explanation drift through its Kendall-based acceptance mechanism.

The separability results further support this conclusion. K-IPO achieved the best or tied-best $1-N3$ value in 13 out of 20 datasets, surpassing SMOTE, which ranked second with 5 wins, and TabDDPM with 3 wins. This suggests that the proposed acceptance mechanism tends to reject synthetic samples that would blur class boundaries or increase the overlap between minority and majority regions. This behavior explains why K-IPO can preserve interpretability without sacrificing predictive performance. By selecting only candidate samples that maintain the original importance structure, this method tends to preserve a more favorable class geometry.

Table~\ref{tab:wins_summary} summarizes the number of wins achieved by each method across all datasets and evaluation metrics. K-IPO clearly dominates the interpretability-related metrics, recording 15 wins for $I$ and 20 for Kendall’s $\tau$, far exceeding all competing methods. It also achieved the strongest class separability, with 13 wins for $1-N3$. K-IPO also leads across the predictive metrics, with 9 wins each in Balanced Accuracy, F1-score, MCC, and $P$. TabDDPM was the closest predictive competitor, achieving 5 wins in each metric, whereas TVAE recorded no wins in any category. Overall, K-IPO is the only method that consistently combines strong predictive performance with explanation consistency, rank preservation, and class separability.

\begin{table}
\centering
\caption{Number of wins achieved by each method across all datasets and selected evaluation metrics.}
\label{tab:wins_summary}
\begin{tabular}{lccccccc}
\hline
\textbf{Method} & \textbf{Bal.\ Acc.} & \textbf{F1} & \textbf{MCC} & $\boldsymbol{P}$ & $\boldsymbol{I}$ & \textbf{Kendall's} $\boldsymbol{\tau}$ & $\boldsymbol{1\!-\!N3}$ \\
\hline
CTGAN            & 4  & 4  & 4  & 4  & 2  & 1  & 1  \\
Gaussian Copula   & 1  & 1  & 1  & 1  & 1  & \underline{3}  & 0  \\
K-IPO            & \textbf{9} & \textbf{9} & \textbf{9} & \textbf{9} & \textbf{15} & \textbf{20} & \textbf{13} \\
SMOTE            & 2  & 2  & 3  & 3  & 1  & 1  & \underline{5}  \\
TabDDPM          & \underline{5}  &  \underline{5}  & \underline{5}  & \underline{5}  & \underline{3}  & 0  & 3  \\
TVAE             & 0  & 0  & 0  & 0  & 0  & 0  & 0  \\
\hline
\end{tabular}
\end{table}

\subsection{Computational Overhead}

\begin{figure}
    \centering
    \includegraphics[width=\linewidth]{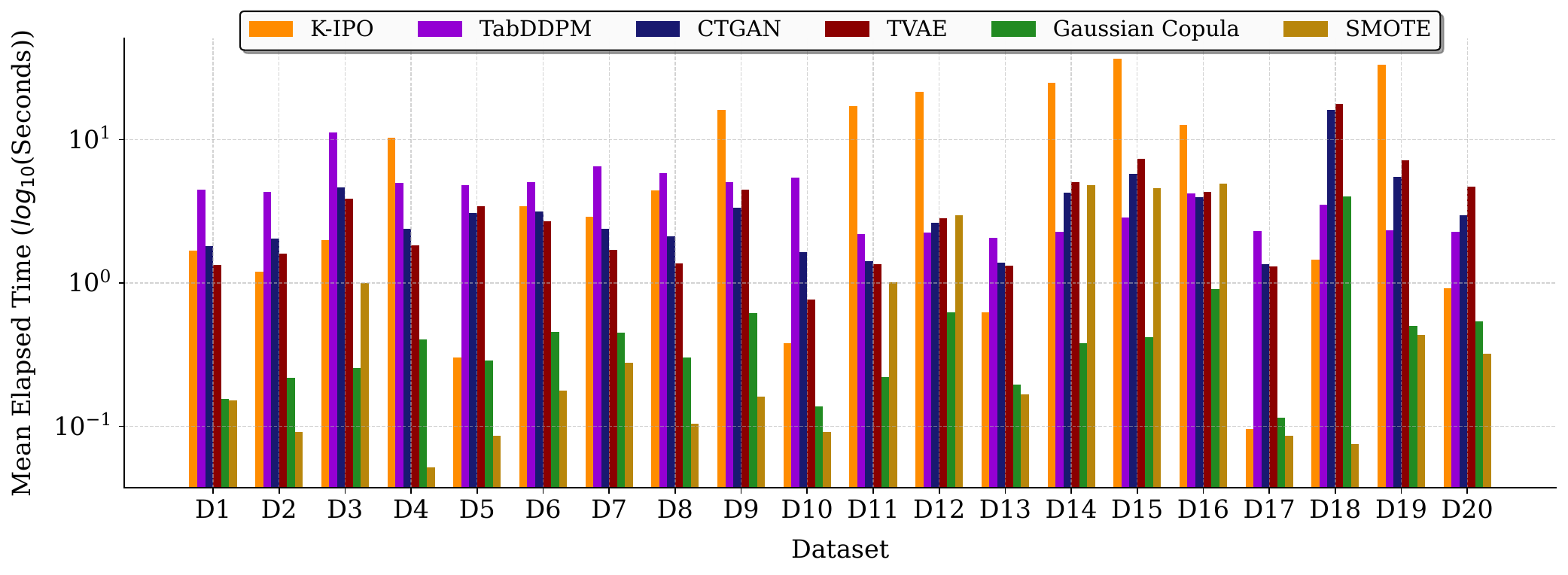}
    \caption{Runtime comparison by dataset and oversampling method on a logarithmic scale.}
    \label{fig:runtime_by_dataset_method_logscale}
\end{figure}

Regarding the computational cost, Fig.~\ref{fig:runtime_by_dataset_method_logscale} shows the mean elapsed time for each dataset and method, while Table~\ref{tab:computational-cost-summary} summarizes the average and median wall-clock times across all datasets. The results revealed clear differences in computational efficiency among method families. Gaussian Copula achieved the lowest overall mean runtime at 0.561 s, followed by SMOTE at 1.082 s. This is expected, as Gaussian Copula relies on a lightweight statistical fit-and-sample procedure, whereas SMOTE generates synthetic minority samples through nearest-neighbor interpolation without training a neural generative model. In contrast, CTGAN, TVAE, and TabDDPM require an explicit model-fitting stage before sampling: CTGAN trains an adversarial network with conditional generation and training-by-sampling, TVAE trains a variational autoencoder over multiple epochs, and TabDDPM trains a diffusion-based generator. Therefore, their runtimes reflect both model learning and sampling costs, which explains their substantially higher computational burden.

K-IPO had the highest mean runtime (9.605 s), primarily due to the repeated feature importance evaluations required by the Kendall-constrained acceptance mechanism. For each candidate block, the resulting feature importance ranking is compared with the reference ranking, and rejected blocks are recursively split into smaller chunks before being accepted or discarded. Thus, K-IPO incurs a computational cost absent from the unconstrained baseline methods. However, the median runtime provides a more nuanced picture. The median runtime of K-IPO was 3.184 s, which was close to those of TVAE (2.768 s) and CTGAN (2.808 s), and lower than that of TabDDPM (4.316 s). Therefore, K-IPO is not consistently slower than neural or diffusion-based generators. Instead, its higher mean runtime is mainly driven by a small number of complex datasets, where the acceptance mechanism becomes highly selective, leading to repeated block rejection and recursive splitting. This pattern is further illustrated in Fig.~\ref{fig:runtime_by_dataset_method_logscale}, where K-IPO exhibited moderate runtimes on several datasets but showed sharp increases in specific cases.

\begin{table}[bp]
\centering
\caption{Summary of computational cost across oversampling methods.}
\label{tab:computational-cost-summary}

\begin{tabular}{lcccc}
\toprule
Method & Mean time (s) & Median time (s) \\
\midrule
SMOTE & 1.082 & 0.173 \\
Gaussian Copula & 0.561 & 0.393 \\
TVAE & 3.816 & 2.768 \\
CTGAN & 3.606 & 2.808 \\
K-IPO & 9.605 & 3.184 \\
TabDDPM & 4.199 & 4.316 \\
\bottomrule
\end{tabular}
\end{table}

Importantly, K-IPO can be faster than deep generative baselines on datasets where valid candidate blocks are accepted early. In such cases, the expensive training phase required by CTGAN, TVAE, and TabDDPM is avoided, with costs mainly limited to local candidate generation and feature importance verification. In contrast, deep generative methods must first fit a model, even when only a small number of synthetic samples are needed to rebalance the minority class. Thus, the relative runtime of K-IPO depends on the strictness of the importance preservation constraint, the quality of the candidate pool, the number of required synthetic samples, and dataset dimensionality. Overall, although K-IPO introduces additional computational costs, this overhead remains acceptable, given its gains in predictive performance, explainability consistency, feature importance rank preservation, and class separability.

\subsection{Statistical Analysis}

To complement the descriptive evaluation, we performed a non-parametric statistical analysis across the examined datasets and competing methods. Following the standard protocol for comparing multiple algorithms over multiple datasets~\cite{Demsar2006StatisticalComparisons}, the methods were ranked within each dataset for each evaluation metric, with lower ranks indicating better performance. We then applied the Friedman test~\cite{Friedman1937RanksANOVA} to assess whether the observed rank differences were statistically significant. When the null hypothesis was rejected, we conducted the Nemenyi post-hoc test~\cite{Nemenyi1963DistributionFreeComparisons} to identify the methods whose average ranks did not differ significantly.

Table~\ref{tab:friedman_results} presents the Friedman test statistics and the corresponding $p$-values. At $\alpha=0.05$, the null hypothesis of equal performance across the baseline methods was rejected for all examined metrics, indicating statistically significant differences among the methods. This pattern holds for predictive metrics, including Balanced Accuracy, F1-score, MCC, and the predictive score $P$, as well as for interpretability-oriented metrics. The strongest effects were observed for explainability consistency $I$, Kendall's $\tau$, and separability $1-N3$. Overall, the results show that the choice of synthetic data generation method significantly influences both predictive performance and interpretability.

\begin{table}
\centering
\caption{Friedman test results for the examined datasets and oversampling methods.}
\label{tab:friedman_results}
\begin{tabular}{lcc}
\toprule
Metric & Friedman statistic & $p$-value \\
\midrule
Balanced Accuracy        & 20.5651 & 1$\times10^{-3}$ \\
F1-score                 & 19.9213 & 1.3$\times10^{-3}$ \\
MCC                      & 17.3605 & 3.9$\times10^{-3}$ \\
$P$    & 14.1702 & 1.46$\times10^{-3}$ \\
$I$   & 27.8674 & 3.86$\times10^{-5}$ \\
Kendall's $\tau$         & 72.4090 & 4.21$\times10^{-14}$ \\
Separability ($1-N3$)    & 56.7024 & 5.82$\times10^{-11}$ \\
\bottomrule
\end{tabular}
\end{table}

\begin{figure*}[t]
\centering
\begin{adjustbox}{max width=\textwidth, max totalheight=0.86\textheight, center}
\begin{minipage}{\textwidth}
\centering

\begin{subfigure}[t]{0.48\textwidth}
    \centering
    \includegraphics[width=\textwidth]{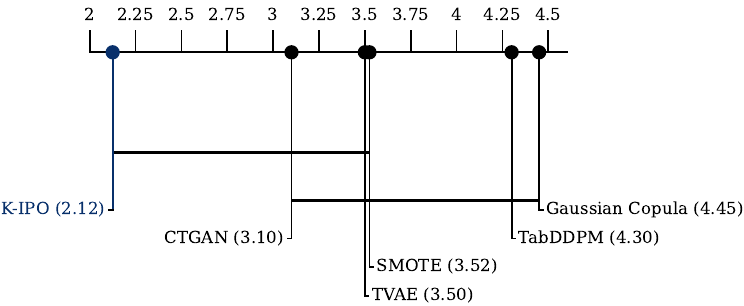}
    \caption{Balanced Accuracy}
\end{subfigure}
\hfill
\begin{subfigure}[t]{0.48\textwidth}
    \centering
    \includegraphics[width=\textwidth]{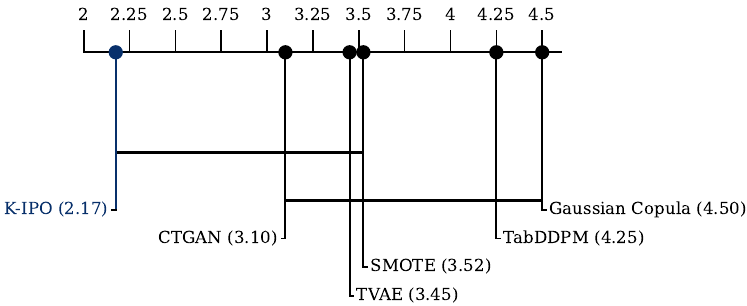}
    \caption{F1-score}
\end{subfigure}

\vspace{0.45em}

\begin{subfigure}[t]{0.48\textwidth}
    \centering
    \includegraphics[width=\textwidth]{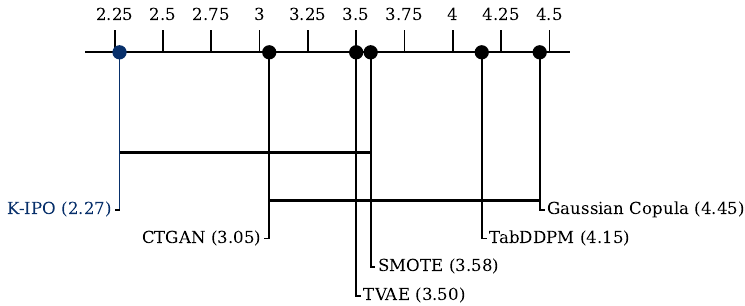}
    \caption{MCC}
\end{subfigure}
\hfill
\begin{subfigure}[t]{0.48\textwidth}
    \centering
    \includegraphics[width=\textwidth]{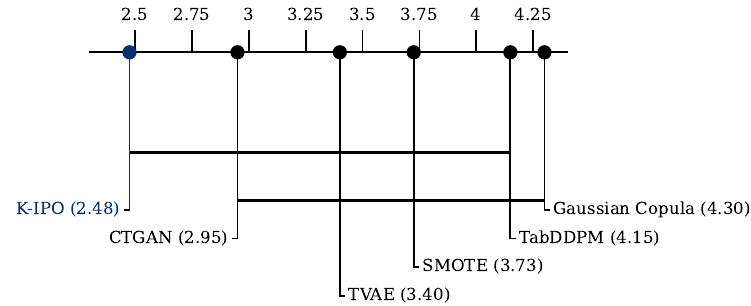}
    \caption{Predictive Score ($P$)}
\end{subfigure}

\vspace{0.45em}

\begin{subfigure}[t]{0.48\textwidth}
    \centering
    \includegraphics[width=\textwidth]{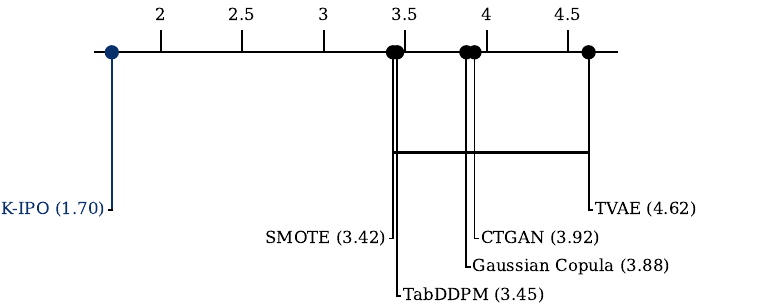}
    \caption{Interpretability ($I$)}
\end{subfigure}
\hfill
\begin{subfigure}[t]{0.48\textwidth}
    \centering
    \includegraphics[width=\textwidth]{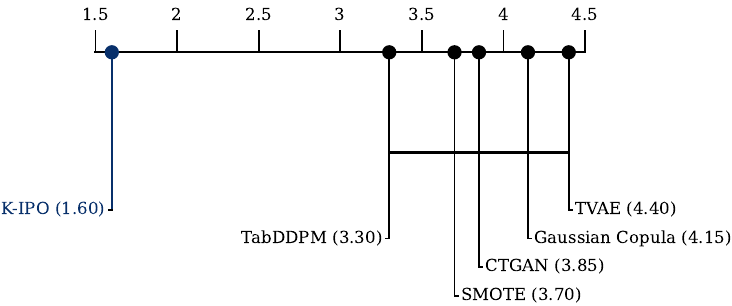}
    \caption{Harmonic Mean $(P,I)$}
\end{subfigure}

\vspace{0.45em}

\begin{subfigure}[t]{0.48\textwidth}
    \centering
    \includegraphics[width=\textwidth]{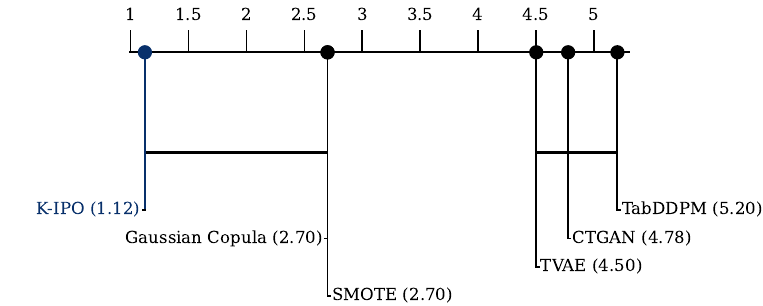}
    \caption{Kendall's $\tau$}
\end{subfigure}
\hfill
\begin{subfigure}[t]{0.48\textwidth}
    \centering
    \includegraphics[width=\textwidth]{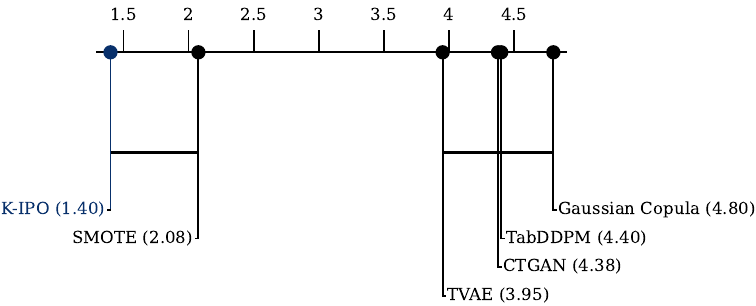}
    \caption{Separability ($1-N3$)}
\end{subfigure}

\end{minipage}
\end{adjustbox}
\caption{Critical difference diagrams from Nemenyi post-hoc tests on predictive, interpretability, and structural metrics. Lower average ranks indicate better performance; methods connected by a horizontal bar are not significantly different at $\alpha = 0.05$.}
\label{fig:cd_diagrams_4x2}
\end{figure*}

The Nemenyi critical difference diagrams in Fig.~\ref{fig:cd_diagrams_4x2} offer a more detailed view of the pairwise post-hoc comparisons. For the predictive metrics, K-IPO achieved the best average rank for Balanced Accuracy, F1-score, MCC, and $P$. However, these differences were not always significant enough to statistically distinguish K-IPO from all competing methods. Specifically, CTGAN, TVAE, SMOTE, and TabDDPM remained statistically comparable to K-IPO for some predictive metrics, indicating that unconstrained oversampling and generative methods can still deliver competitive classification performance on specific datasets.

A clearer pattern emerged for the interpretability-related metrics. For explainability consistency $I$, K-IPO achieved the best average rank by a wide margin (1.70) and formed a distinct top group, demonstrating statistically significant superiority over all competing baselines. The substantial gap from the remaining methods, whose average ranks exceeded 3.4, indicates that the proposed Kendall-constrained acceptance mechanism markedly improved the alignment between feature-attribution explanations and the feature importance ranking of the original data. K-IPO also obtained the best average rank for Kendall’s $\tau$ (1.12), reflecting the strongest preservation of the original feature importance ordering. The post-hoc analysis ranked K-IPO at the top, clearly separating it from lower-performing methods. Although the Gaussian Copula and SMOTE constitute the closest competing group, their average rank (2.70) remains substantially worse, highlighting K-IPO’s superior ability to maintain feature importance consistency. This finding aligns directly with the design objective of K-IPO, where synthetic samples are accepted only if they satisfy an explicit rank-preservation criterion, whereas competing methods do not directly constrain feature importance drift.

The separability results further highlight the benefits of our approach. K-IPO achieved the best average rank for $1-N3$ (1.40) and was statistically comparable only to the SMOTE-based baseline (2.08), while clearly outperforming all other methods, whose average ranks ranged from 3.95 to 4.80. This indicates that the proposed acceptance mechanism effectively filters out synthetic samples that would increase class overlap or distort the local neighborhood structure of the data. As a result, K-IPO preserves class separability more effectively while maintaining the interpretability characteristics of the original dataset.

Finally, the critical difference diagram for the harmonic mean of predictive performance and interpretability provides the strongest overall evidence in favor of K-IPO. K-IPO achieved the best average rank (1.60) and was not connected to any competing method, indicating its statistically significant superiority over all baseline methods. This result shows that K-IPO does not sacrifice interpretability for predictive utility; instead, it offers the most effective overall trade-off between these two objectives. Thus, although some baselines remain competitive in individual predictive metrics, K-IPO is the only method that consistently combines strong predictive performance, explanation consistency, feature importance rank preservation, and class separability.

\section{Conclusions} 
\label{conclusions}

This paper introduced Kendall-constrained Importance-Preserving Oversampling (K-IPO), a generator-agnostic framework for imbalanced tabular classification that explicitly controls feature importance drift during augmentation. Unlike conventional oversampling methods, which focus primarily on class balance, sample plausibility, or predictive performance, K-IPO follows a generate-then-select strategy, accepting candidate minority-class samples only when their inclusion sufficiently preserves the feature importance ranking of the original data. By doing so, K-IPO incorporates explanation stability directly into the augmentation process, rather than treating it solely as a post-hoc evaluation criterion.

Experiments across 20 binary imbalanced datasets, three predictive models, and multiple explanation methods showed that K-IPO achieved the best or tied-best Kendall rank preservation on all examined datasets, the highest explainability-consistency score on 15 datasets, and the strongest class-separability result on 13 datasets. It also remained competitive in predictive performance, recording the most wins in Balanced Accuracy and leading in F1-score, MCC, and the composite predictive score. These findings indicate that explicitly preserving feature importance ranking can improve explanation consistency without systematically compromising classification performance.

Future research should address the computational cost of K-IPO, particularly for large datasets, where repeated feature importance evaluations and candidate subset assessments may limit scalability. Promising directions include parallel implementations for both shared and distributed memory computing environments, faster candidate screening mechanisms, and data-driven methods for the automatic selection of appropriate preservation constraints. The framework could also be extended through uncertainty-aware rank constraints, acceptance criteria that preserve local explanations, and support for multiclass and multilabel classification. Finally, broader evaluations should examine whether the generated samples protect sensitive information, avoid reinforcing unfair biases, preserve causal relationships, and satisfy domain-specific validity requirements.

\bibliographystyle{unsrt}  
\bibliography{references}

\end{document}